%% file: main.tex
\documentclass[accepted]{uai2026} %

\usepackage[american]{babel}

\usepackage{natbib} %
\usepackage{mathtools} %

\usepackage{booktabs} %
\usepackage{tikz} %

\usepackage{times}
\usepackage{amsfonts}
\usepackage{amsmath}
\usepackage{amssymb}
\usepackage{graphicx}
\usepackage{float}
\usepackage{booktabs}
\usepackage{caption}
\usepackage{stfloats}
\usepackage{amsthm}
\usepackage{multirow}
\usepackage{calc}
\usepackage{enumitem}
\usepackage{graphicx} %
\usepackage{relsize}
\usepackage{mathtools}
\usepackage{calc}     %
\usepackage{booktabs}
\usepackage{tabularx}
\usepackage{array}
\usepackage{caption}
\theoremstyle{remark}
\usepackage{cancel}

\usepackage{relsize}
\DeclareMathSizes{10}{10}{8}{9}

\title{Fisher8: Stabilizing Neural Heteroscedastic Regression via Output-Layer Fisher Geometry}

\author[1]{\href{mailto:sumedh.ml@proton.me?Subject=UAI 2026: Fisher8}{Sumedh Vemuganti}}
\author[1]{Nickvash Kani}
\affil[1]{%
    Electrical and Computer Engineering Department\\
    University of Illinois at Urbana-Champaign%
}

\begin{document}
\maketitle
\begin{abstract}
Training neural networks to jointly predict mean and uncertainty estimates from noisy observations can be unstable, prompting a series of independent stabilization efforts. We argue that these interventions highlight a common underlying issue where gradient steps are poorly aligned with the geometry of the loss landscape. To better align updates with local curvature, we derive Fisher8, an output-layer gradient correction that reorients and rescales updates using Fisher geometry rather than Euclidean geometry. Unlike past stabilizers, Fisher8 introduces no data-dependent hyperparameters beyond learning rate and admits an approximate KL trust radius between successive predictive distributions. We show that prior stabilizers converge on overlapping components of this geometric correction. Across multidimensional regression and representation-learning tasks, Fisher8 obtains superior likelihood--error tradeoffs, predicts calibrated uncertainty estimates, and learns rich uncertainty-aware feature spaces. 
\end{abstract}

\input{introduction/intro}

\input{prior_works/prior_works}
\input{methods/methods}

\input{results/results}

\input{prior_works/placing}

\input{conclusion/conclusion}

\bibliography{references/references}    %
\newpage
\onecolumn
\title{Fisher8: Stabilizing Neural Heteroscedastic Regression via Output-Layer Fisher Geometry (Supplemental Material)}
\maketitle
\appendix
\input{appendix/appendix}

\end{document}

%% file: introduction/intro.tex
\section{INTRODUCTION}
Deep learning models that output reliable uncertainty estimates can signal when their predictions should not be trusted (Appendix \ref{app:UQ_Need}). In these settings, models must output both \textit{predictions} and \textit{calibrated uncertainty} estimates, \textit{without seeing uncertainty labels during training} \citep{NixWeigend1994}.

A straightforward approach is to train a neural network to jointly predict a mean and variance using the Gaussian negative log-likelihood (NLL) loss (Eq.  \ref{eq:het-nll}). Despite its conceptual simplicity, this procedure suffers from recurring instabilities in which mean estimates degrade relative to unit-variance baselines~\citep{stirn2023faithful}, models inflate $\sigma^2$ to reduce loss without improving fit~\citep{seitzer2022pitfalls}, and training exhibits sharp phase transitions under varying regularization strengths~\citep{wongtoi2024pathologies}. These failures have motivated selective stop-gradients paired with Newton steps, gradient reweighting, and careful regularization tuning, respectively. While often effective, these stabilizers are largely \emph{prescriptive}: a pathology is observed, and a targeted remedy is engineered.

\emph{If a diverse set of stabilizers is repeatedly required during training, perhaps these interventions are not fixing an unstable loss, but instead correcting a \textbf{misdirected and mis-scaled} traversal of the loss landscape.}

 Fisher8 redirects and rescales output-layer gradients so that updates to the predicted parameters are measured by the approximate KL divergences between the distributions they induce, rather than their Euclidean displacement. We evaluate Fisher8 on regression, parameter prediction, and representation-learning tasks, observing rapid likelihood ascent, reduced regularization dependence, and improved downstream feature space utility. 

We advance uncertainty regression in neural networks by:
\begin{enumerate}[label=\textbf{\arabic*}.,
                  leftmargin=*,
                  itemsep=1pt,
                  topsep=1pt,
                  parsep=0pt,
                  after=\vspace{-\topsep}]
    \item Introducing local KL variance as a diagnostic metric to assess
          when feature-space activity fails to induce distributional change.
    \item Deriving Fisher8, an output-layer natural-gradient correction
          with approximate trust-region control and no dataset-dependent hyperparameters beyond
          learning rate.
    \item Providing a unifying lens that relates independently proposed
          stabilizers to overlapping components of our geometrically motivated
          update rule.
\end{enumerate}

%% file: prior_works/prior_works.tex
\section{\MakeUppercase{The training problem}}
\label{prior:empirical}
Consider a data generating process in which an input  $x_i \in \mathbb{R}^d$ produces a noisy observation $y_i \in \mathbb{R}$.
\begin{equation}
\label{eq:het-dgp}
y_i = \mu(x_i) + \varepsilon_i, \quad \varepsilon_i \sim \mathcal{N}\big(0,\sigma^2(x_i)\big)
\end{equation}
$\mu(x)$ and $\sigma(x)$ are never directly observed while collecting training data, and are unavailable during testing. Given training data lacking uncertainty labels,
\(\textstyle \mathcal{D}_{\text{train}} = \{(x_i, y_i)\}_{i=1}^N\), train a neural network $f^{\mathrm{NN}}_{\Theta}$ with per-point loss
\begin{equation}
\label{eq:het-nll}
\ell_i=\tfrac{1}{2}\hat{\sigma}^{-2}(x_i)(y_i-\hat{\mu}(x_i))^2 + \tfrac{1}{2}\ln\hat{\sigma}^2(x_i),
\end{equation}
that, for each test input \(\textstyle x_{\mathrm{test}}\),
outputs a predicted conditional mean and uncertainty
\(\textstyle (\hat{\mu}(x_{\mathrm{test}}),\, \hat{\sigma}^2(x_{\mathrm{test}}))\). 

\input{prior_works/background_img}

\subsection{PRIOR TRAINING CORRECTIONS}
\label{past}
\textit{Observed Pathology:} \citet{seitzer2022pitfalls} show that when trained under Eq. \ref{eq:het-nll}, the gradient contribution of point $x_i$ is inversely modulated by $\hat{\sigma}(x_i)$.
$\nabla_{\hat{\mu}}\ell \sim \hat{\sigma}^{-2}$,\quad
$\nabla_{\hat{\sigma}}\ell \sim \hat{\sigma}^{-3}$.
Points where the model is already certain (small $\hat{\sigma}(x_i)$) contribute large gradient signals while uncertain regions (large $\hat{\sigma}(x_i)$) are starved of updates.

\textit{Prescribed fix (Fig.~\ref{fig:background}, Col 2):} Rescale the per-point gradient updates by $\hat{\sigma}^{2\beta}(x_i)$, with $\beta$ tuned per dataset to dampen the starvation. When $\beta=0.5$, the inverse modulation is relaxed: $\nabla_{\hat{\mu}}\ell\colon \hat{\sigma}^{-2} \mathbin{\rightarrow} \hat{\sigma}^{-1}$,\quad $\nabla_{\hat{\sigma}}\ell\colon \hat{\sigma}^{-3} \mathbin{\rightarrow} \hat{\sigma}^{-2}$.

\textit{Observed Pathology:} \citet{stirn2023faithful} show that networks trained to jointly predict mean and variance can yield worse mean estimates compared to those trained to predict the mean alone with fixed unit variance.

\textit{Prescribed fix} \textit{(Fig.~\ref{fig:background}, Col 3)}: Decouple the network into a shared trunk with separate mean and variance heads. Block variance gradients from updating the shared trunk. Update the mean head via Newton steps instead of standard first-order gradients.

\textit{Observed Pathology:} \citet{wongtoi2024pathologies} use field theory to abstract network architecture specifics and analyze instability at the function level. They characterize failure regimes where variance absorbs error while the mean is poorly fit, and the data is memorized with no uncertainty predicted.

\textit{Prescribed fix (Fig.~\ref{fig:background}, Col 4):} Augment the objective with one regularizer that balances the likelihood term against the networks' parameters, and another that balances the parameters of the mean and variance networks against each other. Search for regularization coefficients along their derived stability corridor to find a non-degenerate solution.

\subsection{PERSPECTIVE SHIFT}
If reaching a desirable weight configuration requires manual algorithmic interventions (mean warm-up schedules, dataset-specific gradient weighting knobs, design choices about how much output heads share versus separate) and if training only succeeds within a narrow band of carefully tuned hyperparameters, then perhaps the bottleneck is not the specific network architecture or loss function itself, but rather the path we take through the loss landscape. What these interventions show is that the \emph{same} network can reach a good solution once the training dynamics are heavily modified, so the selected model classes are clearly expressive enough. Moreover, once the networks do reach a good solution, the loss is minimized, so the objective function itself is not fundamentally ill-posed, but rather underconstrained. The deeper question is whether our updates are \emph{meaningfully aligned with and constrained by} the geometry of the loss surface, or simply steps in Euclidean space.

%% file: prior_works/background_img.tex
\begin{figure*}[t]
    \captionsetup{labelfont=bf}
    \centering
    \includegraphics[width=1\linewidth]{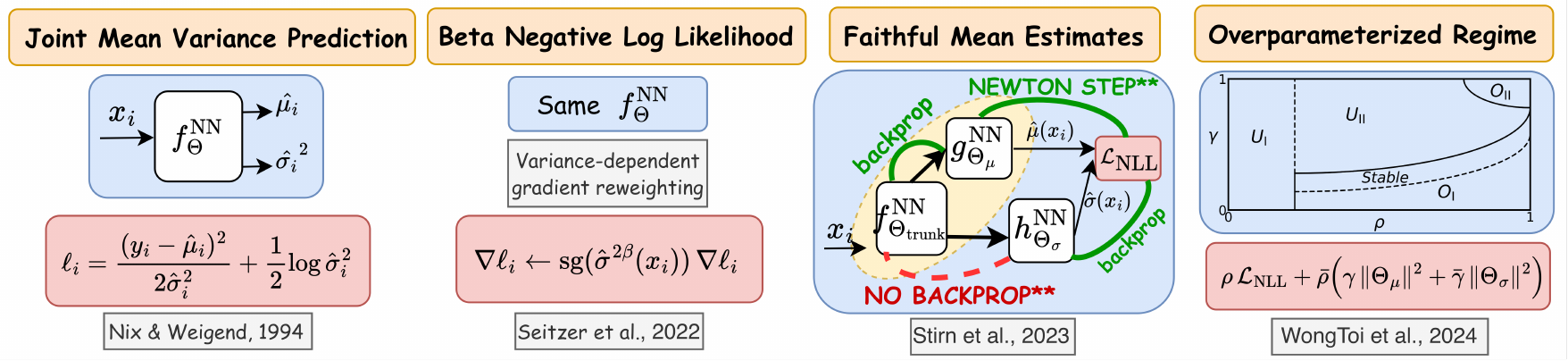}
    \caption{Conceptual summary of four approaches to uncertainty estimation in neural networks. \textbf{Col 1:} Joint mean--variance prediction with Gaussian NLL \citep{NixWeigend1994}. \textbf{Col 2:} Variance-reweighted gradients via Beta-NLL \citep{seitzer2022pitfalls}. \textbf{Col 3:} Faithful mean estimation using a separate MSE head without backpropagation through the shared trunk \citep{stirn2023faithful}. \textbf{Col 4:} Over-parameterization with distinct regularization for mean and variance heads \citep{wongtoi2024pathologies}.}
    \label{fig:background}
\end{figure*}

%% file: methods/methods.tex
\begin{figure*}[t]
        \captionsetup{labelfont=bf}
        \centering
        \includegraphics[width=1\linewidth]{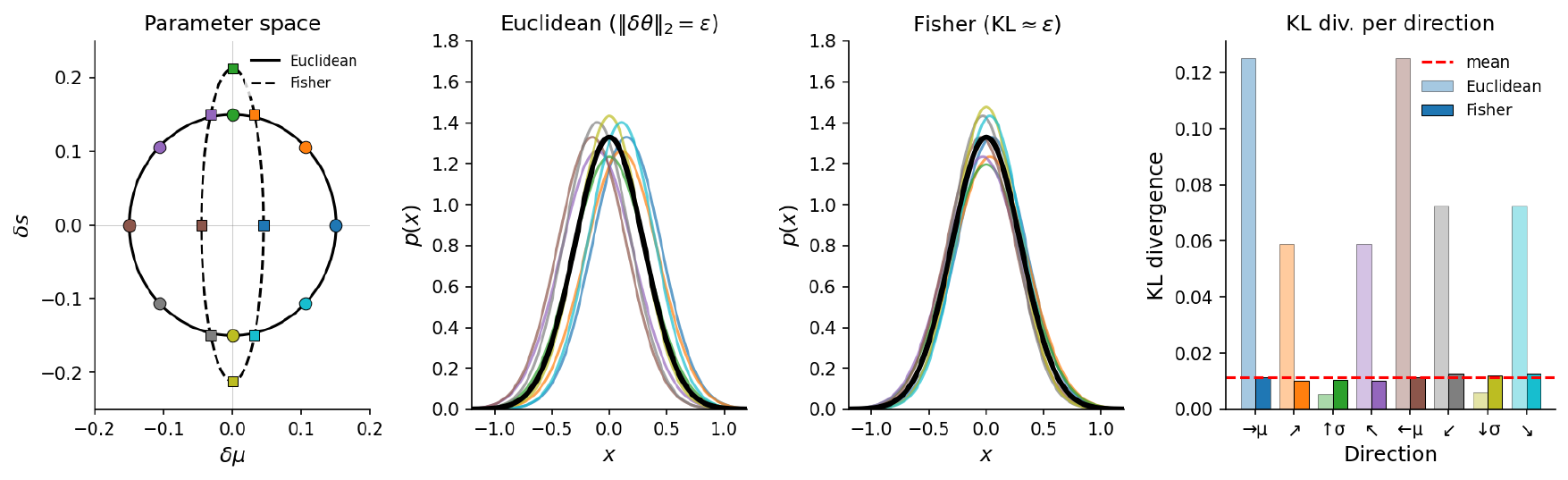}
        \caption{
\textbf{Col 1:} Parameter space showing Euclidean circle (solid) and Fisher ellipse (dashed) with 8 sampled directions.
\textbf{Col 2:} Gaussian PDFs resulting from Euclidean perturbations from the black reference $\mathcal{N}(0,0.3^2)$.
\textbf{Col 3:} Gaussian PDFs resulting from Fisher perturbations.
\textbf{Col 4:} KL divergence from base distribution for each direction under both metrics.}
    \label{fig:kl_fisher_intro}
\end{figure*}

\section{Methodology}
\label{sec:method}
\subsection{Per-Point Update Rule}
\label{methods:direction}
Consider an observation $(x,y)$ generated from Eq. \ref{eq:het-dgp}.
Let $f_\Theta^{\mathrm{NN}}:\mathbb{R}^d \to \mathbb{R}^2$ be a neural network with parameters $\Theta$ that maps an input $x$ to predicted Gaussian distributional parameters $\theta=[\mu,\,s]^\top$ with $s=\ln\sigma^2$. The negative log-probability of observing $y$ yields the loss function
\begin{subequations}\label{eq:loss}
\begin{align}
\ell_{\theta}(y) &= -\ln p_{\theta}(y) + C \label{eq:loss_a}\\
&= \tfrac{1}{2}s + \tfrac{1}{2}e^{-s}(y-\mu)^2, \label{eq:loss_b}
\end{align}
\end{subequations}
with partial derivatives:
\begin{subequations}\label{eq:loss-partials}
\begin{align}
\nabla_{\mu}\ell &= -e^{-s}(y-\mu) \label{eq:loss-partials_a},\\
\nabla_{s}\ell   &= \tfrac{1}{2} - \tfrac{1}{2}e^{-s}(y-\mu)^2,\label{eq:loss-partials_b}\\
\nabla_{\theta}\ell &= [\,\nabla_{\mu}\ell,\ \nabla_{s}\ell\,]^\top. \label{eq:loss-partials_grad}
\end{align}
\end{subequations}

The objective for the steepest descent problem that searches for \textit{the direction} of the parameter updates $\delta\theta=[\delta\mu,\delta s]^\top$ providing the largest reduction in loss $\ell_{\theta}(y)$, as $\varepsilon\to0$, is:
\begin{equation}
\label{eq:steepest}
\begin{aligned}
\arg\min_{\delta\theta}\;
&\ell_{\theta+\delta\theta}
\quad \text{s.t.}\quad
\|\delta\theta\|_2 \le \varepsilon.
\end{aligned}
\end{equation}
Linearizing the loss as $\ell_{\theta+\delta\theta} \approx \ell_{\theta} + \nabla_{\theta}\ell_{\theta}^{\top}\delta\theta$ and taking a step of size $R$ (set by learning rate $\eta$) in the direction of steepest descent, we obtain the gradient descent objective.
\begin{subequations}\label{eq:euclidean-gd-step}
\begin{align}
  & \arg\min_{\delta\theta}\; \ell_{\theta} + \delta\theta^{\top}\nabla_{\theta}\ell_{\theta}
      \quad \text{s.t.}\quad \|\delta\theta\|_{2}\leq R
      \label{eq:euclid_constraint} \\
  & \Longrightarrow\; \delta\theta^* = -\eta\,\nabla_{\theta}\ell_{\theta},
      \quad R = \eta\,\|\nabla_\theta\ell_{\theta}\|_2
      \label{eq:euclid_solution}
\end{align}
\end{subequations}
A close inspection of Eq. \ref{eq:euclid_constraint} reveals an \textit{implicit Euclidean constraint}, forcing parameter perturbations $\delta \theta=(\delta\mu,\delta s)$ to lie on a Euclidean ball $(\delta\mu)^2 + (\delta s)^2 = R^2$, such as the solid black circle in Fig.~\ref{fig:kl_fisher_intro}, Col 1. Each colored point corresponds to a candidate perturbation from $\mathcal{N}(0,0.3^2)$, at the center of the circle, to a new induced distribution $\mathcal{N}(0+\delta\mu,\, 0.3^2e^{\delta s})$ in Fig.~\ref{fig:kl_fisher_intro}, Col 2. These distributions exhibit volatile shifts, relative to the black reference normal. The corresponding translucent bars in Fig.~\ref{fig:kl_fisher_intro}, Col 4 show that the KL divergences of the corresponding perturbed Gaussians have a large spread. \textit{Perturbations in $\mu$ and $s$ along the same, small, fixed Euclidean radius can induce erratic differences in the information separating their Gaussians.}

We now reinterpret $(\mu,s)$ as coordinates on a manifold $\mathcal{M}$,
\begin{equation}
\mathcal{M}
\;\coloneqq\;
\bigl\{\,
\theta = (\mu,s) \in \mathbb{R} \times \mathbb{R}
\;\big|\;
p_\theta = \mathcal{N}\!\left(\mu, e^{\,s}\right)\bigr\},
\label{eq:manifold}
\end{equation}
where distance between coordinates $\theta$ and $\theta+\delta\theta$ is no longer Euclidean, and instead \textit{locally quantifies} the difference between the probability distributions they induce via $\mathrm{KL}\!\left(p_{\theta}\,\|\,p_{\theta+\delta\theta}\right)$ \footnote{KL divergence is not a distance metric as it is asymmetric. When parameters are close, it is approximately locally symmetric.}. The steepest descent \textit{direction search} problem on $\mathcal{M}$ as $\varepsilon\to0$ is 
\begin{equation}
\begin{aligned}
\arg\min_{\delta\theta}\;
&\ell_{\theta+\delta\theta}
\quad \text{s.t.}\quad
\mathrm{KL}\!\left(p_{\theta}\,\|\,p_{\theta+\delta\theta}\right)
\le \varepsilon.
\end{aligned}
\label{eq:ng_opt_setup}
\end{equation}

A second-order Maclaurin\footnote{Not to be confused with the McLaren Formula 1 team, who are allegedly largely uninterested in KL divergences.} series expansion of the constraint from Eq. \ref{eq:ng_opt_setup} \citep{Amari1998} is 
\begin{equation}
\label{eq:kl_fisher_second_order}
\begin{aligned}
\mathrm{KL}&(p_{\theta}\,\|\,p_{\theta+\delta\theta})
=
\mathbb{E}_{p_{\theta}}
\!\left[
\ln p_{\theta}(y)-\ln p_{\theta+\delta\theta}(y)
\right] \\
&\approx
\cancel{\mathrm{KL}(p_{\theta}\,\|\,p_{\theta})}
-\delta\theta^{\top}
\cancel{\mathbb{E}_{p_{\theta}}\!\left[\nabla_{\theta}\ln p_{\theta}(y)\right]} \\
&\qquad{}-\tfrac{1}{2}\delta\theta^{\top}
\mathbb{E}_{p_{\theta}}\!\left[\nabla_{\theta}^{2}\ln p_{\theta}(y)\right]
\delta\theta \\
&=
\tfrac{1}{2}\delta\theta^{\top}\mathcal{F}(\theta)\delta\theta \\
&=
\tfrac{1}{2}\|\mathcal{F}^{1/2}(\theta)\,\delta\theta\|_2^{2},
\end{aligned}
\end{equation}
where $\mathcal{F}(\theta)=-\mathbb{E}_{p_{\theta}}[\nabla_{\theta}^{2}\ln p_{\theta}(y)]$ is the Fisher information matrix (FIM), quantifying the sensitivity of the likelihood to infinitesimal parameter perturbations. Under our parameterization $\theta=(\mu,s)$, $\mathcal{F}(\theta)$ is given as follows:
\begin{subequations}
\label{eq:FIM}
\begin{gather}
\mathcal{F}(\theta) = \operatorname{diag}\!\big(e^{-s},\,\tfrac{1}{2}\big),
\label{eq:FIM-forward} \\
\mathcal{F}(\theta)^{-1} = \operatorname{diag}\!\big(e^{s},\,2\big).
\label{eq:FIM-inverse}
\end{gather}
\end{subequations}

The KL constraint from Eq. \ref{eq:ng_opt_setup} can be approximated with its second-order expansion in Eq. \ref{eq:kl_fisher_second_order}, and FIM from Eq. \ref{eq:FIM-forward}:
\begin{equation}
\label{eq:kl-radius}
\tfrac{1}{2}\delta\theta^{\top}\mathcal F_\theta\,\delta\theta
\;\approx\;
\frac{(\delta\mu)^2}{2\,e^{s}} + \frac{(\delta s)^2}{4} 
\end{equation}

This places the set of feasible parameter perturbations onto a Fisher ellipse, such as the dashed one depicted in Fig.~\ref{fig:kl_fisher_intro}, Col 1. For each colored point, the corresponding perturbed distribution in Fig.~\ref{fig:kl_fisher_intro}, Col 3 undergoes controlled movement. The matching solid, colored bars in Fig.~\ref{fig:kl_fisher_intro}, Col 4 show that the resulting change in Gaussian information is roughly equal across every perturbation.

With first-order loss and second-order KL approximations, the natural gradient descent problem on $\mathcal{M}$ is:
\begin{subequations}
\label{eq:fisher-ngd-step}
\begin{align}
& \arg\min_{\delta\theta}\;
\ell_{\theta}(y) + \delta\theta^{\top}\nabla_{\theta}\ell_{\theta}
\;\text{s.t.}\;
\scalebox{0.87}{$\left\|\mathcal F(\theta)^{1/2}\delta\theta\right\|_{2} \leq P$}
\label{eq:fisher-ngd-constraint} \\
& \Longrightarrow\; \delta\theta^{*} = -\eta\,\mathcal F(\theta)^{-1}\nabla_{\theta}\ell_{\theta}
\eqqcolon -\eta\,\nabla_{\theta}^{\mathrm{nat}}\ell_{\theta}
\label{eq:fisher-ngd-solution} \\
& \phantom{\Longrightarrow\;} \scalebox{0.85}{$P = \eta\left\|\mathcal F(\theta)^{-1/2}\nabla_{\theta}\ell_{\theta}\right\|_{2}$}.
\label{eq:fisher-ngd-radius}
\end{align}
\end{subequations}
The output layer \textit{natural gradients} are:
\begin{subequations}
\setlength{\jot}{3pt}%
\begin{align}
\nabla^{\mathrm{nat}}_{\theta}\ell_{\theta}
&\coloneqq
\mathcal{F}(\theta)^{-1}\nabla_{\theta}\ell_{\theta}
= [\nabla^{\mathrm{nat}}_{\mu}\ell_{\theta},\, \nabla^{\mathrm{nat}}_{s}\ell_{\theta}]^\top
\label{eq:nat-grad-def} \\
\nabla^{\mathrm{nat}}_{\mu}\ell_{\theta}
&= e^{s}\,\nabla_{\mu}\ell_{\theta}
\label{eq:nat-grad-mu} \\
\nabla^{\mathrm{nat}}_{s}\ell_{\theta}
&= 2\,\nabla_{s}\ell_{\theta}.
\label{eq:nat-grad-s}
\end{align}
\end{subequations}

The \textit{per-point} update rule back-propagates the output layer Fisher pre-conditioned gradients through the network's weights.
\begingroup
\setlength{\jot}{0pt}%
\begin{subequations}
\label{eq:perpoint_update_normalized}
\begin{align}
\Theta
&\leftarrow
\Theta
-\eta\,(\nabla_{\theta}^{\mathrm{nat}}\ell_{\theta})^\top
\frac{\partial\theta}{\partial \Theta}
\\
\Theta
&\leftarrow
\Theta
-
\eta
\Big(
e^{s}\,\nabla_{\mu}\ell_{\theta} \; \tfrac{\partial \mu}{\partial \Theta}
+
2\,\nabla_{s}\ell_{\theta} \; \tfrac{\partial s}{\partial \Theta}
\Big)
\label{eq:perpoint_update_normalized-expanded}
\end{align}
\end{subequations}
\endgroup

\subsection{Batched Update Rule}

In practice, however, updates are performed in batches $\textstyle\{(x_i, y_i)\}_{i=1}^B$, where point $x_i$ has its own predicted parameters $\textstyle f_{\Theta}^{NN}(x_i) = \theta_i = (\mu_i, s_i)$ and \textit{its own local Fisher geometry} $\mathcal{F}(\theta_i)$. Under a shared batch learning rate, the distributional distance that each point travels (Eq. \ref{eq:fisher-ngd-constraint}) $\textstyle\tfrac{1}{2}P_i^2$ depends on its local Fisher geometry, and is therefore highly variable: $\textstyle P_i^2 \neq P_j^2$ in general. Using a single learning rate would then attempt to control $B$ different radii with a single knob, \textit{misdirecting traversal}.

On the flip side, normalizing each point's natural gradient by its own local Fisher geometry forces every point in the batch to move the same distributional distance. Different points inherently require different amounts of movement to reach an optimum, so uniform movement \textit{mis-scales traversal}.

\textit{An effective batch update rule must therefore allow the natural variation in intra-batch distributional mobility requirements to persist, while also admitting a reasonable bound on the batch-level trust radius.}

We define the batch loss $\mathcal{L}_{\boldsymbol{\theta}} =
\tfrac{1}{B}\sum_{i=1}^{B}\ell_{\theta_i}(y_i)$. We stack per-point quantities over the batch into bold vectors, where $\boldsymbol{\mu}, \boldsymbol{s}, \nabla_{\boldsymbol{\mu}}, \nabla_{\mathbf{s}} \in \mathbb{R}^{B}$, $\boldsymbol{\theta} \in \mathbb{R}^{B\times 2}$, and subscript $i$ indexes the $i$-th batch sample. The hat operator $\hat{\boldsymbol{v}} = \boldsymbol{v}/\|\boldsymbol{v}\|_2$ normalizes vectors to their unit $L_{2}$ norm and $\odot$ represents element-wise multiplication. The batch update rule computes the per-point Fisher preconditioned output layer gradients, normalizes them to unit $L_{2}$ norm, and back-propagates this augmented signal through the network's weights.
\begin{subequations}
\setlength{\jot}{3pt}%
\label{eq:fisher8_batched_update}
\begin{gather}
\nabla_{\mathbf{s}}^{\mathrm{nat}}
= 2\,\nabla_{\mathbf{s}}
\in \mathbb{R}^{B}
\label{eq:fisher8_natgrad_s}
\\[4pt]
\nabla_{\boldsymbol{\mu}}^{\mathrm{nat}}
= e^{\mathbf{s}} \odot \nabla_{\boldsymbol{\mu}}
\in \mathbb{R}^{B}
\label{eq:fisher8_natgrad_mu}
\\[4pt]
\Theta
\leftarrow
\Theta
-
\eta
\left(
\widehat{\nabla_{\boldsymbol{\mu}}^{\mathrm{nat}}}^{\top}
\,\frac{\partial \boldsymbol{\mu}}{\partial \Theta}
+
\widehat{\nabla_{\mathbf{s}}^{\mathrm{nat}}}^{\top}
\,\frac{\partial \mathbf{s}}{\partial \Theta}
\right)
\label{eq:fisher8_step}
\end{gather}
\end{subequations}

\subsection{Approximate Trust Radius}
The unit normalization in Eq.~\ref{eq:fisher8_step} guarantees $\|\delta\boldsymbol{\mu}\|_2 = \|\delta\mathbf{s}\|_2 = \eta$. \textit{Under nested (1) output-layer-only and (2) independent conditional-likelihood approximations}, a second-order estimate of the batch KL divergence can be measured between the pre- and post-update product distributions.
\begin{equation}
\label{eq:kl_batch_bound}
\begin{aligned}
&\mathrm{KL}\!\left(
\scalebox{0.8}{$\displaystyle\prod$}_{i=1}^{B} p_{\theta_i}
\,\Big\|\,
\scalebox{0.8}{$\displaystyle\prod$}_{i=1}^{B} p_{\theta_i+\delta\theta_i}
\right) \\
&\quad= \sum\nolimits_{i=1}^{B} \mathrm{KL}\!\left(p_{\theta_i}\,\|\,p_{\theta_i+\delta\theta_i}\right) \\
&\quad\approx \tfrac{1}{2}\sum\nolimits_{i=1}^{B} e^{-s_i}(\delta\mu_i)^2 + \tfrac{1}{4}\sum\nolimits_{i=1}^{B}(\delta s_i)^2 \\
&\quad\leq \underbrace{\tfrac{1}{2}\,e^{-\min(\mathbf{s})}\,\eta^2}_{\text{\small KL from \(\delta\boldsymbol{\mu}\)}} + \underbrace{\tfrac{1}{4}\eta^2}_{\text{\small KL from \(\delta\mathbf{s}\)}}
\end{aligned}
\end{equation}

Note that this is not a constraint on the batch optimization problem itself, but rather a post hoc readout of how far the joint output distribution has moved after the update has been applied. 

If prior to the update the network was overconfident (its uncertainties $\mathbf{s}$ were small \textit{despite the mean error being large}), then the distributional contribution from changing the mean (Eq.~\ref{eq:kl_batch_bound}, left under-brace) was large, implying that \textit{a substantial mean correction was applied}. 

Suppose that prior to the update, the network had \textit{inflated its variance estimates} to excuse a large mean error. The KL contribution of further inflating $s$ (Eq.~\ref{eq:kl_batch_bound}, right under-brace) is upper bounded, agnostic of the mean error. With an appropriately set learning rate, the network has \textit{limited incentive to continue to inflate its uncertainty} to mask poor mean fits.

\begin{table*}[t]
\centering
\footnotesize
\setlength{\tabcolsep}{4pt}
\renewcommand{\arraystretch}{1.15}
\resizebox{\textwidth}{!}{%
\begin{tabular}{@{}lcccccc@{}}
\toprule
\textbf{Method} &
\textbf{Mean Update} &
\textbf{Scale Update} &
\textbf{Update Motivation} &
\textbf{KL Control?} &
\textbf{Backward Operator} &
\textbf{Extra Hyperparam?} \\
\midrule
$\beta$-NLL
& $e^{\beta s}\nabla_{\mu}\ell$
& $e^{\beta s}\nabla_{s}\ell$
& Reweight by scale to reduce under-sampling
& \texttimes & stop gradients + reweighting  & $\beta$ \\
Faithful
& $e^{s}\nabla_{\mu}\ell$
& $\nabla_{s}\ell$, severed trunk
& Restore unit-variance RMSE performance
& \texttimes & stop gradients + severing & \texttimes  \\
Fisher8
& $e^{s}\nabla_{\mu}\ell$
& $2\nabla_{s}\ell$
& Change traversal to match loss curvature
& \checkmark & Fisher precondition + norm & \texttimes \\
\bottomrule
\end{tabular}%
}
\caption{Comparison of stabilization mechanisms: update rules, backward-pass operator, and additional hyperparameters.}
\label{tab:stabilizer_summary}
\end{table*}

\subsection{\MakeUppercase{Comparison to Past Stabilizers}}
\label{methods:batched_updates}
As seen in Table~\ref{tab:stabilizer_summary}, Col 2, gradient reweighting that previously required dataset-specific hyperparameter tuning and Newton steps is now directly revealed by the geometry of the loss surface itself. Three independently motivated methods converge on a single MSE update.

While $\beta$-NLL and Fisher8 both scale the log-variance gradients (Col 3), Faithful leaves the scale gradient unaltered, but explicitly decouples its backpropagation into the shared trunk representation as a mechanism for controlling training stability. This procedure is depicted in Fig.~\ref{fig:background}, Col 3. While this restores unit-variance performance, \citet{immer2023effective} document the inadvertent effect of this gradient severing on representation learning. \textit{We show in Section~\ref{res:feature_sapce} that Fisher8 avoids gradient severing, restores performance, and preserves downstream feature-space utility.}

\citet{wongtoi2024pathologies} abstract neural networks with nonparametric twice-differentiable functions. Let \(\Theta_\mu,\Theta_\sigma\) denote the weights of the mean and variance sub-networks and \(\hat{\mu},\hat{\Lambda}\) denote their counterparts in the overparameterized limit. As a network grows infinitely flexible, its parameters become an ill-posed handle for control. Network-norm constraints are no longer enforceable, and are replaced with \textit{penalties on the output-function gradients}.
\begin{subequations}
\label{eq:reg_two_lines}
{\setlength{\jot}{1pt}%
\begin{align}
f^\Theta_{\mathrm{NN}}{:}\quad
&\rho\,\mathcal{L}_{\mathrm{NLL}}
\;+\;
\overbrace{\bar{\rho}\bigl(\gamma\|\Theta_\mu\|^2 + \bar{\gamma}\|\Theta_\sigma\|^2\bigr)}^{\text{weight regularization}}
\label{eq:reg_nn} \\
\widehat{\mu},\widehat{\Lambda}{:}\quad
&\rho\,\mathcal{L}_{\mathrm{NLL}}
\;+\;
\underbrace{\bar{\rho}\bigl(\gamma\|\nabla\widehat{\mu}\|^2 + \bar{\gamma}\|\nabla\widehat{\Lambda}\|^2\bigr)}_{\text{gradient magnitude constraints}}
\label{eq:reg_decoder}
\end{align}%
}
\end{subequations}

Analogously, Fisher8 replaces weight-space penalties with a simple $L_2$ norm constraint on batch gradient magnitudes, motivated by the need to control distributional mobility, rather than weight magnitudes. \textit{We show in Section~\ref{res:feature_sapce} that Fisher8 remains performant without excessive reliance on tuned weight-space regularization terms.}

%% file: results/results.tex
\section{{Experiments}}
\label{sec:experiments}
\input{results/l2_ball_new}
\input{results/UCI_results}
\input{results/cosmology}

\input{results/rotated_mnist}

%% file: results/l2_ball_new.tex
\subsection{1D High-Frequency Sinusoid}
\label{res:sinusoid}
\begin{figure*}[t]
    \captionsetup{labelfont=bf}
    \centering
    \includegraphics[width=1.0\linewidth]{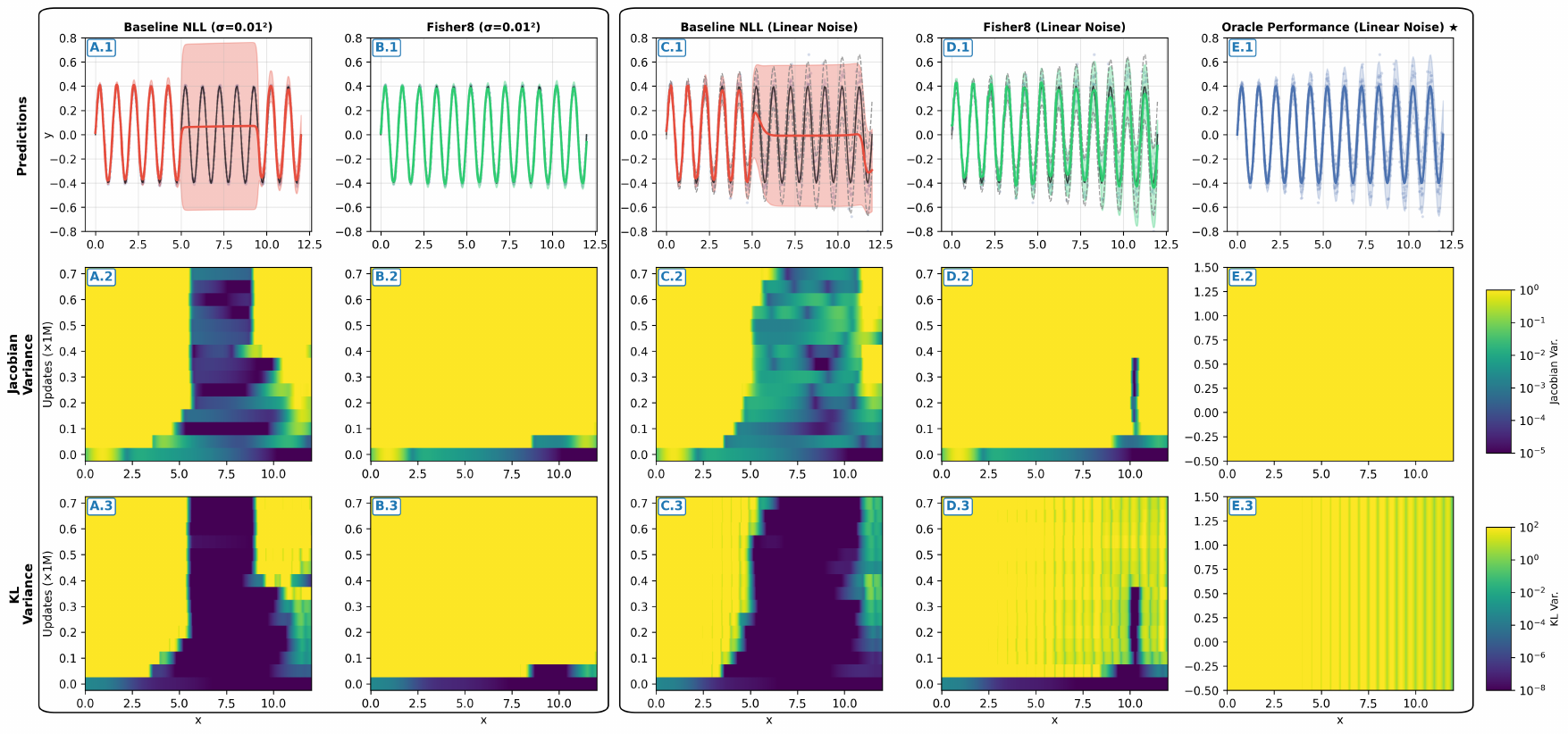}
    \caption{\textbf{Row 1:} predicted fits under input-independent noise (left) and input-dependent noise (right), with the Oracle shown in the rightmost column. Solid red/green lines are the mean predictions of Baseline-NLL/Fisher8; black is the ground-truth mean. Shaded regions show predicted $\pm 2\sigma$; dotted lines show ground-truth $\pm 2\sigma$. \textbf{Row 2:} $\log_{10}$ Jacobian variance of the learned feature space within a 0.5-radius neighborhood over training updates. \textbf{Row 3:} $\log_{10}$ local KL- variance of $\theta=(\mu,s)$ within the same neighborhood over training. The Oracle has high Jacobian variance and displays KL-variance from the ground-truth noise generating functions. We use $r=0.5$ for both diagnostics.}
    \label{fig:visual_fits}
\end{figure*}

Fisher8 corrects the documented training failure on a canonical 1D high-frequency sinusoid with constant noise \citep{seitzer2022pitfalls}. We extend this benchmark to include input-dependent noise and introduce local KL variance as a diagnostic that reveals when feature-space gradient activity fails to produce distributional movement in the output.

\subsubsection{Experimental Setup}
We train two networks (150$\times$150 hidden, tanh, SGD, batch size 32, lr=$10^{-3}$, 700k steps) under Baseline-NLL and Fisher8 on $y = 0.4\sin(2\pi x) + \varepsilon$, under two separate noise regimes: $\sigma(x_i) = 0.01 \;\forall\, x_i$, and $\sigma(x_i) = 0.02+0.01x_i$.
 
 For each model we plot the predicted $\mu$ and $\pm 2\sigma$ against ground truth (Fig.~\ref{fig:visual_fits}, Row 1), the $\log_{10}$ local Jacobian variance~\citep{seitzer2022pitfalls}, measuring how much the Jacobian $J_f$ of the learned features varies within a radius-$r$ $L_2$-ball \textit{as a proxy for feature expressivity/richness} (Fig.~\ref{fig:visual_fits}, Row 2)
\begin{align}
V_J(x) &= \frac{1}{|\mathcal{B}_x|}\sum_{x'\in\mathcal{B}_x}\Bigl\|J_f(x') - \frac{1}{|\mathcal{B}_x|}\sum_{x''\in\mathcal{B}_x}J_f(x'')\Bigr\|^2,\\
\mathcal{B}_x &= \{x'\in\mathcal{D}: \|x-x'\|_2 \leq r\},
\end{align}
and the $\log_{10}$ local KL-variance of the induced output distribution, \textit{as a proxy for the network's ability to support the required distributional mobility} (Fig.~\ref{fig:visual_fits}, Row 3):
\begin{equation}
V_{\mathrm{KL}}(x) = \mathrm{Var}_{x'\in\mathcal{B}_x}\!\Big[\mathrm{KL}\!\big(p_{\theta(x)} \,\|\, p_{\theta(x')}\big)\Big].
\end{equation}

\subsubsection{Results and Analysis}
Plotting conventions referenced in this section are defined in the caption of Fig.~\ref{fig:visual_fits}.

 Columns A and B in Fig.~\ref{fig:visual_fits} compare Baseline-NLL and Fisher8 under input-independent noise. Baseline-NLL (A.1) reproduces the documented pathology in which the predicted mean degrades and the variance inflates to absorb residual error. Fisher8 (B.1) converges to the ground-truth mean with calibrated uncertainty bands within 100k updates, while Baseline-NLL remains stalled for the entirety of the training budget. For Fisher8, Jacobian variance (B.2) and KL-variance (B.3) both develop rich, spatially coupled structure early in training, suggesting that the geometric correction promotes parameter updates that improve both feature-space richness and the KL mobility of the induced output distribution.

Under \textit{input-dependent} noise, Baseline-NLL (C.1) again fails, with the fit degrading most visibly for $x \gtrsim 7$ where the noise is largest. Fisher8 (D.1) \textit{substantially} improves mean tracking and produces variance estimates that follow the linear noise profile, \textit{though a gap relative to the Oracle (E.1) persists at the right end of the domain where Fisher8's noise estimates slightly overshoot at the bottom end of the sinusoid.}

\textit{Where does this remaining error live?} While Jacobian variance (D.2) is uniformly high across the input domain, the KL-variance diagnostic (D.3) retains visible jaggedness between the vertical green bands relative to the Oracle (E.3), \textit{surfacing local imperfections in the predicted distribution that are not captured by feature-space richness alone.} 

In contrast to \textit{local} jaggedness, \textit{global} failure modes in the heatmaps present themselves as \emph{stalling}, in which a region remains uniformly dark across training iterations (A.3), and \emph{churning}, where values are nonzero and visibly fluctuate between iterations (C.2). In the linear noise regime, Baseline-NLL's feature space (C.2) is \textit{churning}, with visible ``activity'' as training progresses, yet KL-variance \textit{stalls}. \textit{Why doesn't the feature space become richer as training progresses despite gradient activity persisting?}

Precisely because activity in the last-layer feature space does not guarantee that the induced output parameters $(\mu,s)$ can vary commensurately across the input domain. While the Jacobian variance (C.2) shows incremental sustained feature-space activity throughout training, the KL-variance (C.3) remains dark; the mapping from features to output parameters compresses incremental gains in richness into near-constant predictions.

We therefore offer an addendum to \citeauthor{seitzer2022pitfalls}'s hypothesis: they argue that early feature-space flatness creates systematic under-sampling and insufficient representational capacity. \textit{We add that incremental gains in feature-space expressivity improve predictive quality only if they also produce sufficient mobility in the distributions induced by the predicted parameters $\theta=(\mu,s)$.}

%% file: results/UCI_results.tex
\subsection{Multidimensional Regression}
\label{experiments:uci} Having evaluated \textit{qualitatively} on \textit{1D} benchmarks, we shift to a \textit{quantitative} evaluation of \textit{multidimensional} UCI~\citep{Dua2019} regression data. Many methods already perform well on these datasets, so we use them not to claim ``state-of-the-art'' performance, but to test whether Fisher8 behaves like a curvature-aware method in practice, despite only being an output-layer intervention. We constrain optimizer settings so that the performance differences are attributed to different methods' gradient interventions, and test whether Fisher8 exhibits empirical fingerprints of second-order methods.

\subsubsection{Experimental Setup}
We use a shared-trunk MLP with 2 hidden layers of 50 units, ELU activations, 2 output heads and train using SGD with batch size 32 for 100 gradient steps.  

The choice of SGD is deliberate. Adam applies per-parameter rescaling that changes both the direction and effective step size of the update, introducing optimizer-level preconditioning. Fisher8 also alters the update direction via an explicit geometry-aware correction at the output layer. Using Adam would therefore superimpose two distinct sources of preconditioning, making it impossible to attribute performance differences specifically to Fisher8. 

While models predict mean and uncertainty estimates, the ground-truth mean and variance functions are unavailable to compare against directly (Section \ref{prior:empirical}). At the same time, models may report artificially high data likelihoods despite having uncalibrated uncertainty estimates and poor mean fits. Robust evaluation therefore requires quantifying quality of mean fit (RMSE), observational likelihood (NLL), and calibration of uncertainty estimates (ECE), where one model's performance is strictly better than another's if it improves on all three statistics. These competing objectives induce a 3-dimensional trade-off space, which we project onto 2D Pareto frontiers in Fig.~\ref{fig:pareto}.  

As a diagnostic for performance at a conservative learning rate, we report statistics for all methods at lr=0.005 in Table~\ref{tab:uci-lr0005}. We then sweep the learning rate across the values in Fig.~\ref{fig:pareto} and track movement across the Pareto frontier.

\input{results/uci_img_tab}

\subsubsection{Results and Analysis}
To test whether Fisher8 updates are geometric in practice, not just in derivation, we assess whether Fisher8 exhibits two documented fingerprints of second-order methods \citep{martens2020new}: (Fingerprint 1) making ``more progress per step'' at conservative learning rates, and (Fingerprint 2) increased sensitivity to learning-rate misspecification when the local approximation is no longer accurate and curvature preconditioning amplifies instability instead of helping.

At learning rate 0.005 with a budget of only 100 gradient steps (Table~\ref{tab:uci-lr0005}), Fisher8 achieves the best RMSE, NLL, and ECE on nearly every benchmark, extracting more progress from each gradient step relative to baselines. The first three columns of Fig.~\ref{fig:pareto} show that Fisher8 dominates the Pareto frontier at all conservative learning rates, consistent with Fingerprint 1.

As the step size exceeds the radius in which the local second-order approximation holds,  we observe a performance reversal of Fisher8 relative to baselines. Fig.~\ref{fig:pareto} shows an asymmetric degradation of Fisher8's Pareto ellipses away from the origin as learning rate increases, while baselines move towards the optimum; large step sizes magnify instability, consistent with Fingerprint 2.

This sweep provides both positive and negative evidence for a geometric explanation of Fisher8's success. If this high-learning-rate degradation were absent, the geometric account would be undermined, and the gains would more likely reflect an artifact other than curvature awareness.

\textit{Limitation: While powerful in its ability to rapidly improve performance, Fisher8 inherits, and is not immune to, the high sensitivity of second-order approaches to the choice of learning rate. It trades the quadratic/cubic hyperparameter search space of baselines (learning rate $\times$ method-specific parameter $\times$ prior regularization) for a finer-resolution search along a single learning-rate axis.}

%% file: results/uci_img_tab.tex
\begin{figure*}
\centering
\captionsetup{type=table,labelfont=bf}
\captionof{table}{UCI benchmark results with SGD (lr=0.005, batch size=32) for 100 steps. Results report mean $\pm$ standard deviation over 20 folds. Bold marks the best (lowest) value per dataset and metric.}
\label{tab:uci-lr0005}
\resizebox{\textwidth}{!}{%
\begin{tabular}{@{}l ccc ccc ccc ccc @{}}
\toprule
& \multicolumn{3}{c}{\textbf{Baseline-NLL}} & \multicolumn{3}{c}{\textbf{$\beta$-NLL (0.5)}} & \multicolumn{3}{c}{\textbf{Faithful}} & \multicolumn{3}{c}{\textbf{Fisher8}} \\
\cmidrule(lr){2-4}\cmidrule(lr){5-7}\cmidrule(lr){8-10}\cmidrule(lr){11-13}
\textbf{Dataset} & \textbf{RMSE}$\downarrow$ & \textbf{NLL}$\downarrow$ & \textbf{ECE}$\downarrow$ & \textbf{RMSE}$\downarrow$ & \textbf{NLL}$\downarrow$ & \textbf{ECE}$\downarrow$ & \textbf{RMSE}$\downarrow$ & \textbf{NLL}$\downarrow$ & \textbf{ECE}$\downarrow$ & \textbf{RMSE}$\downarrow$ & \textbf{NLL}$\downarrow$ & \textbf{ECE}$\downarrow$ \\
\midrule
Yacht & 10.09$\pm$2.72 & 2.50$\pm$0.20 & 0.072$\pm$0.009 & 10.57$\pm$2.54 & 2.85$\pm$0.15 & 0.096$\pm$0.017 & 11.03$\pm$2.51 & 2.78$\pm$0.18 & 0.098$\pm$0.019 & \textbf{8.02$\pm$1.34} & \textbf{2.44$\pm$0.12} & \textbf{0.053$\pm$0.010} \\
Concrete & 11.96$\pm$0.67 & 2.97$\pm$0.05 & 0.024$\pm$0.006 & 12.67$\pm$0.70 & 3.05$\pm$0.05 & 0.028$\pm$0.007 & 13.04$\pm$0.73 & 3.06$\pm$0.06 & \textbf{0.024$\pm$0.006} & \textbf{10.06$\pm$0.54} & \textbf{2.76$\pm$0.06} & 0.026$\pm$0.006 \\
Energy & 3.74$\pm$0.51 & 1.86$\pm$0.09 & 0.065$\pm$0.010 & 4.41$\pm$0.61 & 2.23$\pm$0.05 & 0.075$\pm$0.012 & 4.74$\pm$0.66 & 2.12$\pm$0.09 & 0.060$\pm$0.011 & \textbf{2.99$\pm$0.30} & \textbf{1.37$\pm$0.13} & \textbf{0.035$\pm$0.009} \\
Boston & 5.58$\pm$1.16 & 2.16$\pm$0.11 & 0.051$\pm$0.012 & 6.00$\pm$1.23 & 2.34$\pm$0.11 & 0.062$\pm$0.014 & 6.26$\pm$1.27 & 2.31$\pm$0.13 & 0.054$\pm$0.015 & \textbf{4.44$\pm$1.00} & \textbf{1.93$\pm$0.28} & \textbf{0.035$\pm$0.009} \\
Kin8nm & 0.22$\pm$0.01 & -1.03$\pm$0.04 & 0.013$\pm$0.003 & 0.23$\pm$0.01 & -0.99$\pm$0.04 & \textbf{0.011$\pm$0.003} & 0.23$\pm$0.01 & -0.97$\pm$0.04 & 0.014$\pm$0.004 & \textbf{0.20$\pm$0.00} & \textbf{-1.12$\pm$0.02} & 0.017$\pm$0.003 \\
Naval & 0.01$\pm$0.00 & -4.39$\pm$0.02 & 0.030$\pm$0.004 & 0.01$\pm$0.00 & -4.38$\pm$0.02 & 0.033$\pm$0.004 & 0.01$\pm$0.00 & -4.37$\pm$0.02 & 0.034$\pm$0.004 & \textbf{0.01$\pm$0.00} & \textbf{-4.40$\pm$0.02} & \textbf{0.022$\pm$0.004} \\
Power & 5.47$\pm$0.34 & 2.29$\pm$0.08 & 0.038$\pm$0.007 & 6.11$\pm$0.43 & 2.68$\pm$0.04 & 0.071$\pm$0.006 & 6.43$\pm$0.53 & 2.52$\pm$0.07 & 0.049$\pm$0.007 & \textbf{4.54$\pm$0.14} & \textbf{2.02$\pm$0.04} & \textbf{0.015$\pm$0.004} \\
Wine & 0.70$\pm$0.04 & 0.13$\pm$0.06 & 0.033$\pm$0.008 & 0.72$\pm$0.04 & 0.16$\pm$0.06 & 0.043$\pm$0.012 & 0.73$\pm$0.04 & 0.18$\pm$0.07 & 0.050$\pm$0.011 & \textbf{0.66$\pm$0.04} & \textbf{0.09$\pm$0.11} & \textbf{0.025$\pm$0.006} \\
\bottomrule
\end{tabular}
}
\captionsetup{type=figure,labelfont=bf}
\includegraphics[width=1\linewidth]{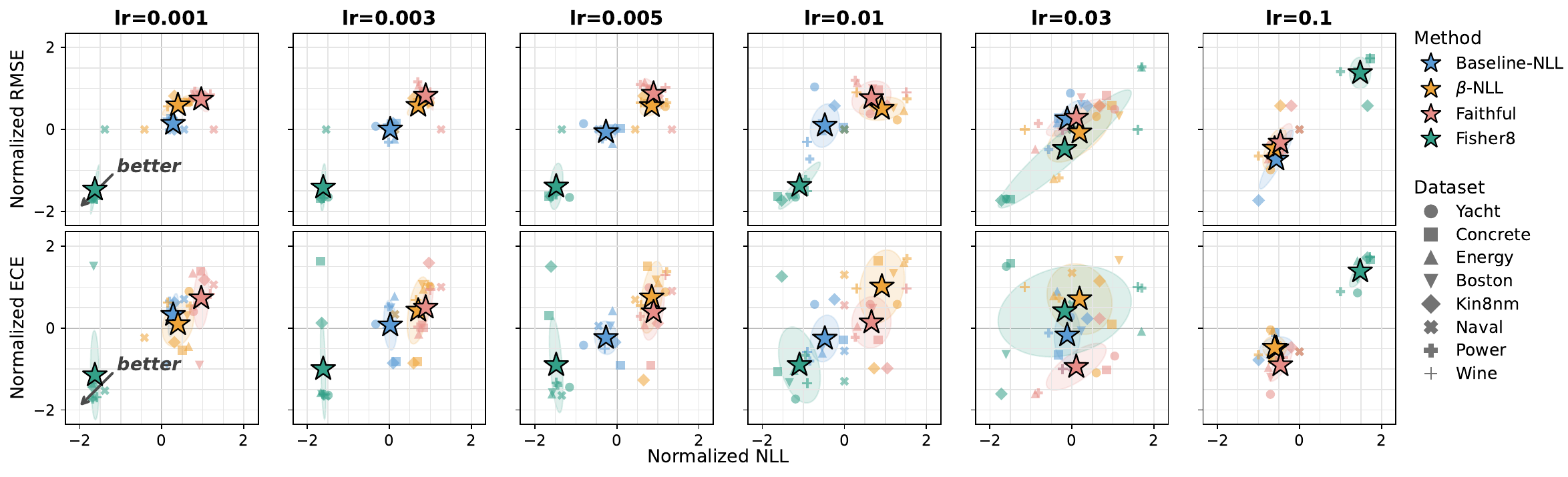}
\caption{Pareto diagrams comparing four heteroscedastic regression methods across eight UCI benchmark datasets at six learning rates (0.001--0.1). Each panel shows normalized RMSE vs.\ normalized NLL, with normalization performed per-dataset using z-scores. Individual dataset results shown as small markers; method means shown as stars with 1-$\sigma$ covariance ellipses; RMSE vs ECE is provided in Fig. \ref{fig:pareto_rmse_ece}. All per-learning-rate tables are provided in Appendix~\ref{app:uci_lr_sweep}.}
\label{fig:pareto}

\end{figure*}

%% file: results/cosmology.tex
\subsection{Weak Lensing Cosmology}

\begin{figure*}[t]
\centering
\captionof{table}{Weak-lensing benchmark. Results report mean $\pm$ standard deviation over 5 initialization seeds. Bold marks the best value per metric (coverage closest to the nominal $0.68$ target).}
\label{tab:weak-lensing}
\resizebox{\textwidth}{!}{%
\begin{tabular}{@{}l ccccccc @{}}
\toprule
\textbf{Method} & \textbf{Score} $\uparrow$ & \textbf{RMSE}$_\Omega\downarrow$ & \textbf{RMSE}$_{S_8}\downarrow$ & \textbf{NLL}$_\Omega\downarrow$ & \textbf{NLL}$_{S_8}\downarrow$ & \textbf{Cov}$_\Omega^{68}$ & \textbf{Cov}$_{S_8}^{68}$ \\
\midrule
Baseline-NLL    & $7.683 \pm 0.261$ & $0.0462 \pm 0.0005$ & $0.0323 \pm 0.0007$ & $-1.651 \pm 0.046$ & $-1.940 \pm 0.075$ & $0.602 \pm 0.054$ & $0.563 \pm 0.045$ \\
$\beta$-NLL (0.5) & $6.952 \pm 0.478$ & $0.0497 \pm 0.0021$ & $0.0332 \pm 0.0005$ & $-1.531 \pm 0.119$ & $-1.897 \pm 0.071$ & $0.564 \pm 0.046$ & $0.559 \pm 0.054$ \\
Faithful       & $6.463 \pm 0.131$ & $0.0539 \pm 0.0013$ & $0.0335 \pm 0.0005$ & $-1.482 \pm 0.024$ & $-1.927 \pm 0.025$ & $0.586 \pm 0.038$ & $\mathbf{0.584 \pm 0.032}$ \\
Fisher8        & $\mathbf{8.528 \pm 0.165}$ & $\mathbf{0.0418 \pm 0.0007}$ & $\mathbf{0.0307 \pm 0.0004}$ & $\mathbf{-1.772 \pm 0.026}$ & $\mathbf{-2.001 \pm 0.067}$ & $\mathbf{0.610 \pm 0.035}$ & $0.576 \pm 0.064$ \\
\bottomrule
\end{tabular}}
\end{figure*}

Weak gravitational lensing refers to the phenomenon where the path of light from a distant galaxy is altered by the gravity of the mass it passes on the way to the telescope, slightly distorting the observed image. These distortions can be converted into an image known as a convergence map $\kappa$, whose structure reflects how much matter the universe holds ($\Omega_m$) and how strongly it has clustered ($S_8$). Fig.~\ref{app:fig_convergence_maps} shows that convergence maps governed by the same underlying parameters can vary widely in appearance. Given convergence maps $\kappa \in \mathbb{R}^{1424 \times 176}$, we wish to infer $(\Omega_m, S_8)$ and uncertainty estimates $(\hat{\sigma}_{\Omega_m}, \hat{\sigma}_{S_8})$.

\subsubsection{Experimental Setup}
 The dataset provided by \citet{fairuniverse2025weaklensing} consists of 101 simulated cosmologies, each observed under 256 noise regimes. We withhold 30\% of realizations across all cosmologies for testing, so models are exposed to all possible cosmological parameters during training, but are evaluated on entirely unseen noise configurations. This provides an extremely \textit{conservative estimate} of performance, and allows us to assess generalizability.

\subsubsection{Results and Analysis}
As seen in Table~\ref{tab:weak-lensing}, Fisher8 achieves the highest score (Eq. \ref{eq:wl_score}), lowest RMSE, and best NLL on both parameters, while coverage remains below nominal for all methods. Surprisingly, $\beta$-NLL and Faithful perform worse than Baseline-NLL. In synthetic benchmarks, uncertainty $\sigma^2(x)$ reflects well-defined observation noise. In addition to Gaussian shape noise, convergence maps carry systematics (baryonic feedback and photometric redshift uncertainty): $\sigma^2(x) = \sigma^2_{\text{noise}}(x) + \sigma^2_{\text{sys}}(x)$. We hypothesize that dataset-dependent gradient reweighting/severing may interact unintentionally with this noise composition, distorting systematics that may, in principle, be learnable through direct training.

%% file: results/rotated_mnist.tex
\subsection{Uncertainty-Aware Representations}
\label{res:feature_sapce}

\subsubsection{Experimental Setup}
We evaluate the quality of Fisher8's feature space on a downstream task that requires uncertainty-aware representations, and demonstrate the reduced need for weight-space regularization. We adopt the experimental setup of \cite{immer2023effective}.

Each MNIST handwritten-digit image $D_i$ of class $c_i \in \{0, \dots, 9\}$ is rotated by an angle $\mathrm{rot}_i$ to produce the network input $x_i$. The regression target $y_i$ is a noisy observation of $\mathrm{rot}_i$. We run two variants where the label noise depends on (Eq.~\ref{eq:rotated_dependent}) and does not depend on (Eq.~\ref{eq:rotated_independent}) the digit class:
\begin{subequations}
\begin{align}
\mathrm{rot_i} &\sim \mathrm{Uniform}(-90^\circ, 90^\circ), \quad \varepsilon_i \sim \mathcal{N}(0,1), \notag \\
y_i &= \mathrm{rot_i} + (11c_i + 1)\,\varepsilon_i, \label{eq:rotated_dependent} \\
y_i &= \mathrm{rot_i} + 10\,\varepsilon_i. \label{eq:rotated_independent}
\end{align}
\end{subequations}
As a \textit{primary representation learning task}, a neural network is trained to predict the rotation angle and its associated uncertainty. As a \textit{secondary downstream task}, the learned feature representations from the penultimate layer are used to train a logistic classifier \textit{to predict the original MNIST digit class}. If the classifier succeeds, the learned representations are rich enough to capture both digit identity and rotation uncertainty, despite the primary training objective being to predict the rotation angle. See Table \ref{tab:mnist_rot}.

\begin{table*}[t]
\centering
\caption{Rotated MNIST under heteroscedastic and homoscedastic noise.
RMSE (rot): rotation angle prediction error; NLL: negative log-likelihood;
ECE: expected calibration error; Feature acc: downstream digit classification
accuracy from learned features. Results report mean $\pm$ standard deviation over 5 initialization seeds. NaturalNLL + KFAC \citep{immer2023effective} is shown as a reference. Bold marks the best value per metric. $\lambda$: prior precision for weight regularization. Additional baselines with method-specific search spaces are reported in Appendix \ref{app:addtl_baselines}.}
\label{tab:mnist_rot}
\resizebox{\textwidth}{!}{%
\begin{tabular}{@{}l cccc cccc @{}}
\toprule
& \multicolumn{4}{c}{\textbf{Input Dependent Noise}}
& \multicolumn{4}{c}{\textbf{Input Independent Noise}} \\
\cmidrule(lr){2-5} \cmidrule(lr){6-9}
\textbf{Method}
& \textbf{RMSE}$^\circ$ $\downarrow$ & \textbf{NLL} $\downarrow$ & \textbf{ECE} $\downarrow$ & \textbf{Feat acc} $\uparrow$
& \textbf{RMSE}$^\circ$ $\downarrow$ & \textbf{NLL} $\downarrow$ & \textbf{ECE} $\downarrow$ & \textbf{Feat acc} $\uparrow$ \\
\midrule
NaturalNLL + KFAC
& $\mathbf{19.492 \pm 1.082}$ & $\mathbf{5.303 \pm 0.018}$ & $0.009 \pm 0.003$ & $76.500 \pm 0.443$
& $\mathbf{12.010 \pm 0.479}$ & $\mathbf{4.085 \pm 0.020}$ & $0.009 \pm 0.004$ & $50.360 \pm 2.366$ \\
\midrule
$\beta$-NLL (0.5)
& $21.488 \pm 2.032$ & $\mathbf{5.363 \pm 0.019}$ & $0.014 \pm 0.006$ & $70.980 \pm 1.714$
& $14.869 \pm 1.921$ & $4.256 \pm 0.084$ & $0.028 \pm 0.012$ & $51.800 \pm 4.713$ \\
$\beta$-NLL (1.0)
& $22.138 \pm 2.935$ & $5.398 \pm 0.048$ & $0.015 \pm 0.011$ & $67.440 \pm 2.010$
& $\mathbf{13.980 \pm 1.500}$ & $\mathbf{4.244 \pm 0.083}$ & $0.019 \pm 0.016$ & $35.440 \pm 3.220$ \\
Faithful
& $21.925 \pm 1.171$ & $5.555 \pm 0.013$ & $0.024 \pm 0.001$ & $54.220 \pm 0.952$
& $14.507 \pm 0.695$ & $4.293 \pm 0.047$ & $0.016 \pm 0.010$ & $37.360 \pm 1.628$ \\
Fisher8 ($\lambda=0.1$)
& $\mathbf{20.934 \pm 0.339}$ & $5.387 \pm 0.011$ & $\mathbf{0.005 \pm 0.001}$ & $\mathbf{82.120 \pm 0.382}$
& $14.708 \pm 0.246$ & $4.252 \pm 0.013$ & $\mathbf{0.006 \pm 0.001}$ & $80.940 \pm 0.320$ \\
Fisher8 ($\lambda=0$)
& $\mathbf{20.934 \pm 0.339}$ & $5.387 \pm 0.011$ & $\mathbf{0.005 \pm 0.001}$ & $82.080 \pm 0.319$
& $14.709 \pm 0.246$ & $4.252 \pm 0.013$ & $\mathbf{0.006 \pm 0.001}$ & $\mathbf{80.980 \pm 0.337}$ \\
\bottomrule
\end{tabular}}
\end{table*}

\subsubsection{Results and Analysis}
All methods perform grid search over prior precision $\lambda$ for weight regularization during cross-validation. Baselines select $\lambda=100$ or $1000$ via cross-validation, while Fisher8 selects $\lambda=0.1$, a reduction of at least $1000\times$ in regularization. Further, completely disabling this regularization (last row) creates negligible changes in Fisher8's performance. \textit{This gap suggests that much of the burden of hyperparameter search over weight-space regularization can be alleviated through constraints on output layer gradient magnitudes}.

While Faithful attains reasonable errors and likelihoods in the input-dependent case, its downstream accuracy collapses to 54.2\%. \textit{We suspect that blocking the gradients originating at the variance head from flowing into the common backbone ($\Theta_{\mathrm{trunk}}$ in Fig.~\ref{fig:background}, Col 3) may be severing critical information flow into the shared representation, inhibiting utility of the learned feature space on downstream tasks.} 

Most strikingly, Fisher8 obtains a classifier accuracy of 80\% in the input-independent noise regime, while other methods collapse to 35--52\% (Table \ref{tab:mnist_rot}, Col 8). Recall that the primary objective is simply to predict the rotation angle. With class-independent noise there is no need to learn a feature space that also encodes digit identity, yet Fisher8 learns digit-aware features by default. \textit{The definition of distance in Fisher8's updates is rooted in the amount of information separating the current and candidate probability distributions that explain the data.}

%% file: prior_works/placing.tex
\section{Relation to Prior Works}
\label{sec:scope}

\textbf{Class 1: Single Pass Direct Mean--Variance Prediction.}
These works were discussed extensively in Section~\ref{past}. \textit{We place Fisher8 here and reinterpret how these stabilizers overlap with curvature awareness.}

\textbf{Class 2: Ensembles.}
MC Dropout~\citep{Gal2016DropoutAsBayesian} estimates uncertainty from stochastic forward passes with dropout active at test time. While framed as Bayesian, it is often viewed as an implicit ensemble of subnetworks. \citet{lakshminarayanan2017simple} train independently initialized networks and decompose model uncertainty from signal noise. \textit{Fisher8 is a drop-in training modification compatible with either approach.}

\textbf{Class 3: Bayesian Neural Networks.} BNNs infer a posterior over weights, either over the full network or only the last layer~\citep{watson2021latent,kristiadi2020being}, and evaluate $p(y\mid x,\mathcal{D})$ via variational inference~\citep{graves2011practical,blundell2015weight} or Laplace approximations~\citep{daxberger2021laplace}. \textit{Fisher8 does not introduce new methods for posterior inference or test-time sampling.} 

\textbf{Class 4: Natural Parameters.} Scaling natural gradients to deep networks requires weight-space Fisher approximations such as KFAC~\citep{martens2015optimizing}. \citet{megerle2023stable} instead obtain natural gradients of the predictive Gaussian via local parameterizations with per-input trust-region projections, and \citet{immer2023effective} use natural parameters for a better-conditioned objective. \textit{Fisher8 avoids loss surface reparameterization and weight-space curvature entirely, applying the exact $2\times2$ inverse Fisher information matrix only to the output layer gradients.}

%% file: conclusion/conclusion.tex
\section{Conclusion}
We have examined several stabilizers commonly used in neural uncertainty regression and argue that their repeated necessity reflects a misdirected loss surface traversal. We derived an output-layer update rule that measures progress by KL divergence and showed when prior prescriptions mimic components of a Fisher-corrected and distributionally controlled output-layer step. Empirically, Fisher8 exhibits the learning-rate sensitivity signature of natural-gradient methods and learns stronger uncertainty-aware features while requiring less weight-space regularization. Taken together, these results suggest that many independently proposed stabilizers may address overlapping facets of the same underlying geometric issue. Making this connection explicit should enable simpler training of heteroscedastic neural networks. 

\section{Future Work}
Future work includes the extension of Fisher8 to other likelihood classes, a detailed comparison of our output layer intervention to methods that incorporate the shared network curvature, a deeper analysis of the interaction between Adam's preconditioning and Fisher8's gradient reorientation, and expansion from single- or dual-parameter prediction to dense prediction tasks, such as depth estimation, where per-pixel distances and uncertainties are needed.

%% file: appendix/appendix.tex
\input{appendix/UCI_Supplement/uci_app}

\input{appendix/weak_lensing_details}
\input{appendix/rotated_mnist}

\input{appendix/name_meaning}
\input{appendix/wild_berries}

%% file: appendix/UCI_Supplement/uci_app.tex
\section{UCI Experiment Details}

\subsection{Additional Training Details}
Each network was trained and evaluated using the predefined splits of \citet{hernandez2015probabilistic}. Experiments have a fixed budget of 100 updates, no early stopping, no further guard rails such as $\sigma$ clamping, and batch size of 32, with $0.5\log 2\pi$ omitted from all reported likelihoods.

\subsection{Complete Learning Rate Sweep}
\label{app:uci_lr_sweep}
\input{appendix/UCI_Supplement/uci_lr_sgd}

\subsection{ECE-RMSE Pareto Frontier for UCI Across Learning Rates}
 Errors, likelihoods, and calibration metrics induce a 3D tradeoff space. Two projections are visualized in Section~\ref{experiments:uci}. The missing projection follows the same trend and is provided here for completeness.
 
 \begin{figure}[H]
\captionsetup{type=figure,labelfont=bf}
\includegraphics[width=1\linewidth]{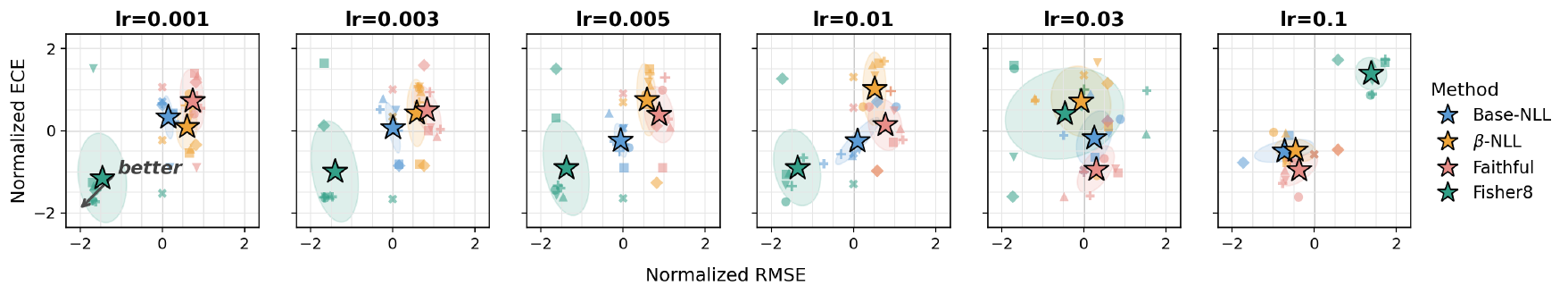}
\caption{Pareto diagrams comparing four heteroscedastic regression methods across eight UCI benchmark datasets at six learning rates (0.001--0.1). Each panel shows normalized ECE vs.\ normalized RMSE, per-dataset. Individual dataset results shown as small markers; method means shown as stars with 1-$\sigma$ covariance ellipses.}
\label{fig:pareto_rmse_ece}
\end{figure}

\subsection{Adam Instead of SGD}
The relationship between Adam and natural gradients is nuanced, beyond the Gaussian regression setting. Some sources characterize Adam as employing a diagonal empirical Fisher preconditioner~\citep{kingma2015adam, hwang2024fadam}, with the latter arguing this grounding holds primarily for discrete distributions, while \citet{kunstner2019limitations} show that the empirical Fisher does not capture second-order curvature in general. This is why we originally scoped our comparison to SGD. The main experiments use SGD to isolate Fisher8's output-layer correction from the per-parameter preconditioning that Adam introduces. Here we retrain all methods with Adam for 100 epochs at batch size 32 at lr=0.001, lr=0.01, and lr=0.1. 

\input{appendix/UCI_Supplement/uci_lr_adam}

%% file: appendix/UCI_Supplement/uci_lr_sgd.tex
\begin{center}
\centering
\captionsetup{type=table,labelfont=bf}
\scriptsize
\setlength{\tabcolsep}{3pt}
\renewcommand{\arraystretch}{1.0}
\captionof{table}{UCI benchmark results with SGD (lr=0.001, batch size=32) for 100 steps. Results report mean $\pm$ standard deviation over 20 folds. Bold marks the best (lowest) value per dataset and metric.}
\label{tab:uci-lr0001}

\resizebox{\textwidth}{!}{%
\begin{tabular}{@{}l ccc ccc ccc ccc ccc @{}}
\toprule
& \multicolumn{3}{c}{\textbf{Unit Variance}} & \multicolumn{3}{c}{\textbf{Baseline-NLL}} & \multicolumn{3}{c}{\textbf{$\beta$-NLL (0.5)}} & \multicolumn{3}{c}{\textbf{Faithful}} & \multicolumn{3}{c}{\textbf{Fisher8}} \\
\cmidrule(lr){2-4}\cmidrule(lr){5-7}\cmidrule(lr){8-10}\cmidrule(lr){11-13}\cmidrule(lr){14-16}
\textbf{Dataset} & \textbf{RMSE}$\downarrow$ & \textbf{NLL}$\downarrow$ & \textbf{ECE}$\downarrow$ & \textbf{RMSE}$\downarrow$ & \textbf{NLL}$\downarrow$ & \textbf{ECE}$\downarrow$ & \textbf{RMSE}$\downarrow$ & \textbf{NLL}$\downarrow$ & \textbf{ECE}$\downarrow$ & \textbf{RMSE}$\downarrow$ & \textbf{NLL}$\downarrow$ & \textbf{ECE}$\downarrow$ & \textbf{RMSE}$\downarrow$ & \textbf{NLL}$\downarrow$ & \textbf{ECE}$\downarrow$ \\
\midrule
Yacht & 13.95$\pm$2.81 & 3.17$\pm$0.17 & 0.100$\pm$0.020 & 13.14$\pm$2.97 & 3.03$\pm$0.24 & 0.087$\pm$0.020 & 13.76$\pm$2.83 & 3.14$\pm$0.22 & 0.091$\pm$0.019 & 13.95$\pm$2.81 & 3.17$\pm$0.26 & 0.087$\pm$0.018 & \textbf{9.95$\pm$2.46} & \textbf{2.49$\pm$0.20} & \textbf{0.070$\pm$0.017} \\
Concrete & 15.48$\pm$0.93 & 3.25$\pm$0.05 & 0.024$\pm$0.006 & 14.85$\pm$0.91 & 3.21$\pm$0.08 & 0.030$\pm$0.005 & 15.33$\pm$0.92 & 3.24$\pm$0.07 & 0.026$\pm$0.006 & 15.48$\pm$0.93 & 3.29$\pm$0.10 & 0.034$\pm$0.007 & \textbf{12.43$\pm$0.75} & \textbf{3.00$\pm$0.06} & \textbf{0.023$\pm$0.005} \\
Energy & 8.00$\pm$0.92 & 2.63$\pm$0.07 & 0.046$\pm$0.009 & 6.76$\pm$0.94 & 2.42$\pm$0.12 & 0.047$\pm$0.008 & 7.71$\pm$0.91 & 2.56$\pm$0.10 & 0.046$\pm$0.009 & 8.00$\pm$0.92 & 2.60$\pm$0.14 & 0.048$\pm$0.011 & \textbf{3.70$\pm$0.53} & \textbf{1.63$\pm$0.14} & \textbf{0.045$\pm$0.014} \\
Boston & 8.42$\pm$1.20 & 2.65$\pm$0.12 & 0.049$\pm$0.014 & 7.72$\pm$1.14 & 2.54$\pm$0.17 & 0.045$\pm$0.012 & 8.24$\pm$1.18 & 2.64$\pm$0.17 & 0.046$\pm$0.012 & 8.42$\pm$1.20 & 2.70$\pm$0.24 & \textbf{0.045$\pm$0.010} & \textbf{5.71$\pm$1.24} & \textbf{2.18$\pm$0.18} & 0.047$\pm$0.013 \\
Kin8nm & 0.26$\pm$0.01 & -0.86$\pm$0.04 & \textbf{0.010$\pm$0.003} & 0.25$\pm$0.01 & -0.84$\pm$0.06 & 0.028$\pm$0.005 & 0.26$\pm$0.01 & -0.84$\pm$0.05 & 0.021$\pm$0.005 & 0.26$\pm$0.01 & -0.78$\pm$0.07 & 0.032$\pm$0.006 & \textbf{0.23$\pm$0.01} & \textbf{-1.00$\pm$0.04} & 0.013$\pm$0.003 \\
Naval & 0.01$\pm$0.00 & -4.39$\pm$0.02 & \textbf{0.023$\pm$0.003} & 0.01$\pm$0.00 & -4.31$\pm$0.05 & 0.045$\pm$0.006 & 0.01$\pm$0.00 & -4.35$\pm$0.03 & 0.037$\pm$0.005 & 0.01$\pm$0.00 & -4.28$\pm$0.06 & 0.048$\pm$0.006 & \textbf{0.01$\pm$0.00} & \textbf{-4.39$\pm$0.02} & 0.026$\pm$0.004 \\
Power & 12.97$\pm$1.02 & 3.13$\pm$0.05 & 0.035$\pm$0.004 & 10.19$\pm$1.24 & 2.85$\pm$0.11 & 0.031$\pm$0.005 & 12.28$\pm$1.04 & 3.03$\pm$0.07 & 0.030$\pm$0.005 & 12.97$\pm$1.02 & 3.07$\pm$0.09 & 0.030$\pm$0.007 & \textbf{5.44$\pm$0.32} & \textbf{2.18$\pm$0.06} & \textbf{0.011$\pm$0.002} \\
Wine & 0.80$\pm$0.04 & 0.28$\pm$0.05 & 0.087$\pm$0.012 & 0.78$\pm$0.04 & 0.31$\pm$0.09 & 0.077$\pm$0.017 & 0.79$\pm$0.04 & 0.30$\pm$0.08 & 0.084$\pm$0.015 & 0.80$\pm$0.04 & 0.39$\pm$0.11 & 0.088$\pm$0.015 & \textbf{0.71$\pm$0.04} & \textbf{0.15$\pm$0.06} & \textbf{0.042$\pm$0.010} \\
\bottomrule
\end{tabular}%
}

\captionof{table}{UCI benchmark results with SGD (lr=0.003, batch size=32) for 100 steps. Results report mean $\pm$ standard deviation over 20 folds. Bold marks the best (lowest) value per dataset and metric.}
\label{tab:uci-lr0003}

\resizebox{\textwidth}{!}{%
\begin{tabular}{@{}l ccc ccc ccc ccc ccc @{}}
\toprule
& \multicolumn{3}{c}{\textbf{Unit Variance}} & \multicolumn{3}{c}{\textbf{Baseline-NLL}} & \multicolumn{3}{c}{\textbf{$\beta$-NLL (0.5)}} & \multicolumn{3}{c}{\textbf{Faithful}} & \multicolumn{3}{c}{\textbf{Fisher8}} \\
\cmidrule(lr){2-4}\cmidrule(lr){5-7}\cmidrule(lr){8-10}\cmidrule(lr){11-13}\cmidrule(lr){14-16}
\textbf{Dataset} & \textbf{RMSE}$\downarrow$ & \textbf{NLL}$\downarrow$ & \textbf{ECE}$\downarrow$ & \textbf{RMSE}$\downarrow$ & \textbf{NLL}$\downarrow$ & \textbf{ECE}$\downarrow$ & \textbf{RMSE}$\downarrow$ & \textbf{NLL}$\downarrow$ & \textbf{ECE}$\downarrow$ & \textbf{RMSE}$\downarrow$ & \textbf{NLL}$\downarrow$ & \textbf{ECE}$\downarrow$ & \textbf{RMSE}$\downarrow$ & \textbf{NLL}$\downarrow$ & \textbf{ECE}$\downarrow$ \\
\midrule
Yacht & 12.27$\pm$2.67 & 3.07$\pm$0.14 & 0.109$\pm$0.022 & 11.04$\pm$2.98 & 2.72$\pm$0.22 & 0.089$\pm$0.016 & 11.88$\pm$2.72 & 2.96$\pm$0.18 & 0.106$\pm$0.023 & 12.27$\pm$2.67 & 2.93$\pm$0.20 & 0.097$\pm$0.019 & \textbf{8.52$\pm$1.41} & \textbf{2.51$\pm$0.11} & \textbf{0.057$\pm$0.014} \\
Concrete & 14.09$\pm$0.82 & 3.17$\pm$0.04 & 0.028$\pm$0.007 & 13.07$\pm$0.81 & 3.05$\pm$0.06 & \textbf{0.023$\pm$0.006} & 13.78$\pm$0.81 & 3.12$\pm$0.06 & 0.023$\pm$0.006 & 14.09$\pm$0.82 & 3.14$\pm$0.07 & 0.024$\pm$0.005 & \textbf{10.49$\pm$0.54} & \textbf{2.80$\pm$0.05} & 0.026$\pm$0.008 \\
Energy & 5.80$\pm$0.80 & 2.48$\pm$0.04 & 0.064$\pm$0.013 & 4.45$\pm$0.67 & 2.08$\pm$0.09 & 0.062$\pm$0.010 & 5.39$\pm$0.76 & 2.33$\pm$0.06 & 0.064$\pm$0.012 & 5.80$\pm$0.80 & 2.29$\pm$0.10 & 0.051$\pm$0.010 & \textbf{3.13$\pm$0.35} & \textbf{1.44$\pm$0.12} & \textbf{0.034$\pm$0.005} \\
Boston & 7.04$\pm$1.25 & 2.52$\pm$0.10 & 0.062$\pm$0.015 & 6.24$\pm$1.18 & 2.28$\pm$0.12 & 0.053$\pm$0.013 & 6.78$\pm$1.22 & 2.43$\pm$0.13 & 0.057$\pm$0.015 & 7.04$\pm$1.25 & 2.43$\pm$0.15 & 0.050$\pm$0.014 & \textbf{4.62$\pm$1.04} & \textbf{1.95$\pm$0.23} & \textbf{0.038$\pm$0.010} \\
Kin8nm & 0.24$\pm$0.01 & -0.91$\pm$0.04 & \textbf{0.011$\pm$0.003} & 0.23$\pm$0.01 & -0.97$\pm$0.05 & 0.015$\pm$0.004 & 0.24$\pm$0.01 & -0.93$\pm$0.05 & 0.015$\pm$0.004 & 0.24$\pm$0.01 & -0.90$\pm$0.05 & 0.020$\pm$0.005 & \textbf{0.20$\pm$0.00} & \textbf{-1.11$\pm$0.03} & 0.017$\pm$0.003 \\
Naval & 0.01$\pm$0.00 & -4.39$\pm$0.01 & \textbf{0.023$\pm$0.003} & 0.01$\pm$0.00 & -4.37$\pm$0.03 & 0.035$\pm$0.004 & 0.01$\pm$0.00 & -4.37$\pm$0.03 & 0.035$\pm$0.005 & 0.01$\pm$0.00 & -4.35$\pm$0.03 & 0.039$\pm$0.005 & \textbf{0.01$\pm$0.00} & \textbf{-4.40$\pm$0.01} & 0.023$\pm$0.004 \\
Power & 7.98$\pm$0.85 & 2.95$\pm$0.02 & 0.070$\pm$0.009 & 6.13$\pm$0.46 & 2.49$\pm$0.08 & 0.052$\pm$0.006 & 7.30$\pm$0.73 & 2.76$\pm$0.05 & 0.062$\pm$0.008 & 7.98$\pm$0.85 & 2.69$\pm$0.08 & 0.043$\pm$0.008 & \textbf{4.67$\pm$0.17} & \textbf{2.04$\pm$0.04} & \textbf{0.013$\pm$0.003} \\
Wine & 0.76$\pm$0.04 & 0.23$\pm$0.05 & 0.073$\pm$0.015 & 0.73$\pm$0.04 & 0.18$\pm$0.07 & 0.051$\pm$0.013 & 0.75$\pm$0.04 & 0.22$\pm$0.07 & 0.062$\pm$0.011 & 0.76$\pm$0.04 & 0.24$\pm$0.08 & 0.066$\pm$0.011 & \textbf{0.66$\pm$0.04} & \textbf{0.09$\pm$0.10} & \textbf{0.025$\pm$0.007} \\
\bottomrule
\end{tabular}%
}

\captionof{table}{UCI benchmark results with SGD (lr=0.01, batch size=32) for 100 steps. Results report mean $\pm$ standard deviation over 20 folds. Bold marks the best (lowest) value per dataset and metric.}
\label{tab:uci-lr001}

\resizebox{\textwidth}{!}{%
\begin{tabular}{@{}l ccc ccc ccc ccc ccc @{}}
\toprule
& \multicolumn{3}{c}{\textbf{Unit Variance}} & \multicolumn{3}{c}{\textbf{Baseline-NLL}} & \multicolumn{3}{c}{\textbf{$\beta$-NLL (0.5)}} & \multicolumn{3}{c}{\textbf{Faithful}} & \multicolumn{3}{c}{\textbf{Fisher8}} \\
\cmidrule(lr){2-4}\cmidrule(lr){5-7}\cmidrule(lr){8-10}\cmidrule(lr){11-13}\cmidrule(lr){14-16}
\textbf{Dataset} & \textbf{RMSE}$\downarrow$ & \textbf{NLL}$\downarrow$ & \textbf{ECE}$\downarrow$ & \textbf{RMSE}$\downarrow$ & \textbf{NLL}$\downarrow$ & \textbf{ECE}$\downarrow$ & \textbf{RMSE}$\downarrow$ & \textbf{NLL}$\downarrow$ & \textbf{ECE}$\downarrow$ & \textbf{RMSE}$\downarrow$ & \textbf{NLL}$\downarrow$ & \textbf{ECE}$\downarrow$ & \textbf{RMSE}$\downarrow$ & \textbf{NLL}$\downarrow$ & \textbf{ECE}$\downarrow$ \\
\midrule
Yacht & 9.37$\pm$1.97 & 2.92$\pm$0.07 & 0.083$\pm$0.014 & 10.38$\pm$2.98 & 2.18$\pm$0.23 & 0.064$\pm$0.013 & 9.14$\pm$1.92 & 2.70$\pm$0.11 & 0.064$\pm$0.012 & 9.37$\pm$1.97 & 2.53$\pm$0.15 & 0.064$\pm$0.013 & \textbf{6.25$\pm$1.32} & \textbf{2.06$\pm$0.22} & \textbf{0.046$\pm$0.012} \\
Concrete & 11.49$\pm$0.52 & 3.05$\pm$0.02 & 0.040$\pm$0.006 & 10.71$\pm$0.54 & 2.85$\pm$0.04 & 0.025$\pm$0.007 & 11.19$\pm$0.51 & 2.94$\pm$0.03 & 0.030$\pm$0.005 & 11.49$\pm$0.52 & 2.94$\pm$0.04 & 0.025$\pm$0.007 & \textbf{9.25$\pm$0.56} & \textbf{2.68$\pm$0.07} & \textbf{0.023$\pm$0.007} \\
Energy & 3.78$\pm$0.50 & 2.38$\pm$0.02 & 0.105$\pm$0.017 & 3.47$\pm$0.47 & 1.60$\pm$0.18 & 0.047$\pm$0.011 & 3.60$\pm$0.46 & 2.09$\pm$0.04 & 0.084$\pm$0.011 & 3.78$\pm$0.50 & 1.81$\pm$0.10 & 0.058$\pm$0.011 & \textbf{3.06$\pm$0.29} & \textbf{1.43$\pm$0.19} & \textbf{0.040$\pm$0.010} \\
Boston & 5.36$\pm$1.18 & 2.40$\pm$0.07 & 0.077$\pm$0.018 & 4.98$\pm$1.07 & 1.98$\pm$0.13 & 0.045$\pm$0.010 & 5.15$\pm$1.13 & 2.22$\pm$0.09 & 0.064$\pm$0.013 & 5.36$\pm$1.18 & 2.15$\pm$0.13 & 0.055$\pm$0.014 & \textbf{4.18$\pm$0.97} & \textbf{1.87$\pm$0.31} & \textbf{0.036$\pm$0.009} \\
Kin8nm & 0.21$\pm$0.01 & -1.00$\pm$0.02 & 0.020$\pm$0.005 & 0.21$\pm$0.01 & -1.09$\pm$0.03 & 0.016$\pm$0.003 & 0.21$\pm$0.01 & -1.06$\pm$0.03 & 0.013$\pm$0.003 & 0.21$\pm$0.01 & -1.05$\pm$0.03 & \textbf{0.013$\pm$0.003} & \textbf{0.20$\pm$0.00} & \textbf{-1.13$\pm$0.02} & 0.017$\pm$0.003 \\
Naval & 0.01$\pm$0.00 & \textbf{-4.40$\pm$0.01} & \textbf{0.022$\pm$0.003} & 0.01$\pm$0.00 & -4.39$\pm$0.02 & 0.025$\pm$0.004 & 0.01$\pm$0.00 & -4.39$\pm$0.02 & 0.030$\pm$0.004 & 0.01$\pm$0.00 & -4.39$\pm$0.02 & 0.028$\pm$0.004 & \textbf{0.01$\pm$0.00} & -4.39$\pm$0.02 & 0.023$\pm$0.004 \\
Power & 5.64$\pm$0.32 & 2.89$\pm$0.01 & 0.100$\pm$0.005 & 4.79$\pm$0.23 & 2.06$\pm$0.04 & \textbf{0.016$\pm$0.005} & 5.44$\pm$0.29 & 2.53$\pm$0.05 & 0.068$\pm$0.004 & 5.64$\pm$0.32 & 2.28$\pm$0.05 & 0.028$\pm$0.005 & \textbf{4.57$\pm$0.17} & \textbf{2.04$\pm$0.05} & 0.019$\pm$0.007 \\
Wine & 0.68$\pm$0.04 & 0.15$\pm$0.04 & 0.035$\pm$0.007 & 0.67$\pm$0.04 & 0.09$\pm$0.07 & 0.027$\pm$0.006 & 0.68$\pm$0.04 & 0.10$\pm$0.06 & 0.029$\pm$0.007 & 0.68$\pm$0.04 & 0.11$\pm$0.06 & 0.029$\pm$0.008 & \textbf{0.66$\pm$0.04} & \textbf{0.09$\pm$0.12} & \textbf{0.026$\pm$0.006} \\
\bottomrule
\end{tabular}%
}

\captionof{table}{UCI benchmark results with SGD (lr=0.03, batch size=32) for 100 steps. Results report mean $\pm$ standard deviation over 20 folds. Bold marks the best (lowest) value per dataset and metric.}
\label{tab:uci-lr003}

\resizebox{\textwidth}{!}{%
\begin{tabular}{@{}l ccc ccc ccc ccc ccc @{}}
\toprule
& \multicolumn{3}{c}{\textbf{Unit Variance}} & \multicolumn{3}{c}{\textbf{Baseline-NLL}} & \multicolumn{3}{c}{\textbf{$\beta$-NLL (0.5)}} & \multicolumn{3}{c}{\textbf{Faithful}} & \multicolumn{3}{c}{\textbf{Fisher8}} \\
\cmidrule(lr){2-4}\cmidrule(lr){5-7}\cmidrule(lr){8-10}\cmidrule(lr){11-13}\cmidrule(lr){14-16}
\textbf{Dataset} & \textbf{RMSE}$\downarrow$ & \textbf{NLL}$\downarrow$ & \textbf{ECE}$\downarrow$ & \textbf{RMSE}$\downarrow$ & \textbf{NLL}$\downarrow$ & \textbf{ECE}$\downarrow$ & \textbf{RMSE}$\downarrow$ & \textbf{NLL}$\downarrow$ & \textbf{ECE}$\downarrow$ & \textbf{RMSE}$\downarrow$ & \textbf{NLL}$\downarrow$ & \textbf{ECE}$\downarrow$ & \textbf{RMSE}$\downarrow$ & \textbf{NLL}$\downarrow$ & \textbf{ECE}$\downarrow$ \\
\midrule
Yacht & 8.68$\pm$1.36 & 2.89$\pm$0.04 & 0.071$\pm$0.012 & 9.24$\pm$2.85 & 2.26$\pm$0.35 & 0.063$\pm$0.015 & 8.43$\pm$1.75 & 2.43$\pm$0.12 & \textbf{0.053$\pm$0.011} & 8.68$\pm$1.36 & 2.55$\pm$0.10 & 0.056$\pm$0.013 & \textbf{5.56$\pm$1.72} & \textbf{1.84$\pm$0.40} & 0.072$\pm$0.027 \\
Concrete & 10.31$\pm$0.54 & 3.01$\pm$0.02 & 0.049$\pm$0.005 & 9.92$\pm$0.67 & 2.70$\pm$0.07 & 0.025$\pm$0.006 & 10.14$\pm$0.54 & 2.80$\pm$0.04 & 0.027$\pm$0.006 & 10.31$\pm$0.54 & 2.79$\pm$0.05 & \textbf{0.024$\pm$0.007} & \textbf{8.52$\pm$0.69} & \textbf{2.61$\pm$0.08} & 0.031$\pm$0.008 \\
Energy & 3.26$\pm$0.37 & 2.36$\pm$0.01 & 0.116$\pm$0.014 & 3.35$\pm$0.50 & 1.70$\pm$0.25 & 0.056$\pm$0.014 & \textbf{3.16$\pm$0.35} & 1.67$\pm$0.07 & 0.055$\pm$0.010 & 3.26$\pm$0.37 & \textbf{1.51$\pm$0.11} & \textbf{0.033$\pm$0.007} & 3.54$\pm$0.66 & 2.43$\pm$3.39 & 0.047$\pm$0.018 \\
Boston & 4.63$\pm$1.03 & 2.35$\pm$0.05 & 0.083$\pm$0.014 & 4.60$\pm$1.07 & 1.95$\pm$0.52 & 0.046$\pm$0.015 & 4.55$\pm$1.01 & 1.99$\pm$0.10 & 0.052$\pm$0.012 & 4.63$\pm$1.03 & 1.95$\pm$0.17 & \textbf{0.043$\pm$0.011} & \textbf{4.18$\pm$1.12} & \textbf{1.87$\pm$0.26} & 0.044$\pm$0.011 \\
Kin8nm & 0.20$\pm$0.00 & -1.04$\pm$0.01 & 0.029$\pm$0.003 & 0.20$\pm$0.00 & -1.13$\pm$0.02 & 0.017$\pm$0.003 & 0.20$\pm$0.00 & -1.12$\pm$0.02 & 0.018$\pm$0.003 & 0.20$\pm$0.00 & -1.12$\pm$0.02 & 0.017$\pm$0.003 & \textbf{0.19$\pm$0.01} & \textbf{-1.20$\pm$0.06} & \textbf{0.015$\pm$0.004} \\
Naval & 0.01$\pm$0.00 & \textbf{-4.40$\pm$0.02} & 0.022$\pm$0.003 & 0.01$\pm$0.00 & -4.40$\pm$0.01 & \textbf{0.022$\pm$0.004} & 0.01$\pm$0.00 & -4.40$\pm$0.02 & 0.025$\pm$0.003 & 0.01$\pm$0.00 & -4.40$\pm$0.02 & 0.022$\pm$0.004 & \textbf{0.01$\pm$0.00} & -4.40$\pm$0.02 & 0.024$\pm$0.005 \\
Power & 4.92$\pm$0.22 & 2.88$\pm$0.00 & 0.110$\pm$0.003 & 4.81$\pm$0.38 & 2.09$\pm$0.07 & 0.022$\pm$0.010 & \textbf{4.69$\pm$0.18} & 2.10$\pm$0.04 & 0.029$\pm$0.006 & 4.92$\pm$0.22 & \textbf{2.08$\pm$0.04} & \textbf{0.010$\pm$0.002} & 5.16$\pm$0.77 & 2.18$\pm$0.16 & 0.031$\pm$0.013 \\
Wine & \textbf{0.66$\pm$0.04} & 0.12$\pm$0.04 & 0.037$\pm$0.007 & 0.66$\pm$0.04 & 0.08$\pm$0.11 & \textbf{0.026$\pm$0.007} & 0.66$\pm$0.04 & \textbf{0.07$\pm$0.06} & 0.027$\pm$0.006 & 0.66$\pm$0.04 & 0.08$\pm$0.07 & 0.026$\pm$0.006 & 0.66$\pm$0.04 & 0.10$\pm$0.15 & 0.027$\pm$0.007 \\
\bottomrule
\end{tabular}%
}

\captionof{table}{UCI benchmark results with SGD (lr=0.1, batch size=32) for 100 steps. Results report mean $\pm$ standard deviation over 20 folds. Bold marks the best (lowest) value per dataset and metric. \texttt{----} denotes a metric that diverged.}
\label{tab:uci-lr01}

\resizebox{\textwidth}{!}{%
\begin{tabular}{@{}l ccc ccc ccc ccc ccc @{}}
\toprule
& \multicolumn{3}{c}{\textbf{Unit Variance}} & \multicolumn{3}{c}{\textbf{Baseline-NLL}} & \multicolumn{3}{c}{\textbf{$\beta$-NLL (0.5)}} & \multicolumn{3}{c}{\textbf{Faithful}} & \multicolumn{3}{c}{\textbf{Fisher8}} \\
\cmidrule(lr){2-4}\cmidrule(lr){5-7}\cmidrule(lr){8-10}\cmidrule(lr){11-13}\cmidrule(lr){14-16}
\textbf{Dataset} & \textbf{RMSE}$\downarrow$ & \textbf{NLL}$\downarrow$ & \textbf{ECE}$\downarrow$ & \textbf{RMSE}$\downarrow$ & \textbf{NLL}$\downarrow$ & \textbf{ECE}$\downarrow$ & \textbf{RMSE}$\downarrow$ & \textbf{NLL}$\downarrow$ & \textbf{ECE}$\downarrow$ & \textbf{RMSE}$\downarrow$ & \textbf{NLL}$\downarrow$ & \textbf{ECE}$\downarrow$ & \textbf{RMSE}$\downarrow$ & \textbf{NLL}$\downarrow$ & \textbf{ECE}$\downarrow$ \\
\midrule
Yacht & 7.44$\pm$1.33 & 2.85$\pm$0.04 & 0.080$\pm$0.018 & --- & --- & 0.093$\pm$0.192 & \textbf{6.28$\pm$2.73} & 2.37$\pm$1.28 & 0.078$\pm$0.030 & 7.44$\pm$1.33 & \textbf{2.34$\pm$0.14} & \textbf{0.050$\pm$0.015} & 10.80$\pm$5.09 & 30.45$\pm$108.52 & 0.094$\pm$0.033 \\
Concrete & 9.01$\pm$0.56 & 2.96$\pm$0.02 & 0.060$\pm$0.007 & 8.80$\pm$1.07 & 2.64$\pm$0.12 & 0.031$\pm$0.009 & \textbf{8.73$\pm$0.63} & \textbf{2.63$\pm$0.07} & 0.025$\pm$0.006 & 9.01$\pm$0.56 & 2.66$\pm$0.07 & \textbf{0.025$\pm$0.006} & 11.53$\pm$2.03 & 3.06$\pm$0.36 & 0.047$\pm$0.014 \\
Energy & \textbf{2.98$\pm$0.30} & 2.35$\pm$0.01 & 0.118$\pm$0.015 & --- & --- & 0.040$\pm$0.033 & 3.03$\pm$0.38 & 1.52$\pm$0.13 & 0.051$\pm$0.010 & 2.98$\pm$0.30 & \textbf{1.37$\pm$0.10} & \textbf{0.033$\pm$0.007} & 7.77$\pm$2.59 & 4.48$\pm$4.87 & 0.085$\pm$0.034 \\
Boston & 4.25$\pm$0.94 & 2.33$\pm$0.05 & 0.087$\pm$0.011 & --- & --- & 0.056$\pm$0.018 & \textbf{4.24$\pm$0.95} & \textbf{1.88$\pm$0.34} & 0.043$\pm$0.017 & 4.25$\pm$0.94 & 1.89$\pm$0.24 & \textbf{0.039$\pm$0.009} & 6.99$\pm$1.56 & 3.52$\pm$1.80 & 0.060$\pm$0.014 \\
Kin8nm & 0.20$\pm$0.01 & -1.06$\pm$0.01 & 0.031$\pm$0.003 & \textbf{0.19$\pm$0.01} & \textbf{-1.17$\pm$0.04} & \textbf{0.016$\pm$0.003} & 0.20$\pm$0.01 & -1.15$\pm$0.03 & 0.017$\pm$0.002 & 0.20$\pm$0.01 & -1.14$\pm$0.03 & 0.017$\pm$0.003 & 0.20$\pm$0.02 & -1.07$\pm$0.13 & 0.024$\pm$0.012 \\
Naval & 0.01$\pm$0.00 & \textbf{-4.40$\pm$0.02} & 0.022$\pm$0.003 & \textbf{0.01$\pm$0.00} & -4.40$\pm$0.02 & \textbf{0.022$\pm$0.004} & 0.01$\pm$0.00 & -4.40$\pm$0.02 & 0.022$\pm$0.003 & 0.01$\pm$0.00 & -4.40$\pm$0.02 & 0.022$\pm$0.003 & --- & --- & 0.025$\pm$0.010 \\
Power & \textbf{4.52$\pm$0.13} & 2.87$\pm$0.00 & 0.115$\pm$0.002 & --- & --- & 0.069$\pm$0.192 & 4.78$\pm$0.58 & \textbf{2.09$\pm$0.11} & 0.029$\pm$0.010 & 4.52$\pm$0.13 & --- & \textbf{0.014$\pm$0.009} & 9.41$\pm$2.73 & 3.75$\pm$3.36 & 0.065$\pm$0.021 \\
Wine & \textbf{0.66$\pm$0.04} & 0.12$\pm$0.04 & 0.035$\pm$0.006 & 0.66$\pm$0.04 & 0.09$\pm$0.13 & 0.027$\pm$0.007 & 0.66$\pm$0.04 & \textbf{0.09$\pm$0.11} & 0.027$\pm$0.006 & 0.66$\pm$0.04 & 0.10$\pm$0.15 & \textbf{0.026$\pm$0.005} & 0.70$\pm$0.05 & 0.18$\pm$0.13 & 0.037$\pm$0.011 \\
\bottomrule
\end{tabular}%
}
\end{center}

%% file: appendix/UCI_Supplement/uci_lr_adam.tex
\begin{center}
\captionsetup{type=table,labelfont=bf}
\captionof{table}{UCI benchmark results with Adam, epoch mode (100 epochs) (lr=0.001, batch size=32). Results report mean $\pm$ standard deviation over 20 folds.}
\label{tab:uci-adam-epoch-lr0001}
\resizebox{\textwidth}{!}{%
\begin{tabular}{@{}l ccc ccc ccc ccc ccc@{}}
\toprule
& \multicolumn{3}{c}{\textbf{Unit Variance}} & \multicolumn{3}{c}{\textbf{Baseline-NLL}} & \multicolumn{3}{c}{\textbf{$\beta$-NLL (0.5)}} & \multicolumn{3}{c}{\textbf{Faithful}} & \multicolumn{3}{c}{\textbf{Fisher8}} \\
\cmidrule(lr){2-4}\cmidrule(lr){5-7}\cmidrule(lr){8-10}\cmidrule(lr){11-13}\cmidrule(lr){14-16}
\textbf{Dataset} & \textbf{RMSE}$\downarrow$ & \textbf{NLL}$\downarrow$ & \textbf{ECE}$\downarrow$ & \textbf{RMSE}$\downarrow$ & \textbf{NLL}$\downarrow$ & \textbf{ECE}$\downarrow$ & \textbf{RMSE}$\downarrow$ & \textbf{NLL}$\downarrow$ & \textbf{ECE}$\downarrow$ & \textbf{RMSE}$\downarrow$ & \textbf{NLL}$\downarrow$ & \textbf{ECE}$\downarrow$ & \textbf{RMSE}$\downarrow$ & \textbf{NLL}$\downarrow$ & \textbf{ECE}$\downarrow$ \\
\midrule
Yacht & 1.52$\pm$0.35 & 2.72$\pm$0.02 & 0.291$\pm$0.068 & 8.85$\pm$2.76 & 0.02$\pm$0.47 & 0.059$\pm$0.014 & 0.79$\pm$0.29 & -0.31$\pm$0.26 & 0.049$\pm$0.014 & 1.52$\pm$0.36 & 0.63$\pm$0.15 & 0.046$\pm$0.011 & 1.51$\pm$0.43 & 0.84$\pm$0.30 & 0.050$\pm$0.014 \\
Concrete & 5.73$\pm$0.53 & 2.88$\pm$0.01 & 0.098$\pm$0.009 & 5.87$\pm$0.58 & 2.21$\pm$0.23 & 0.026$\pm$0.008 & 5.73$\pm$0.54 & 2.20$\pm$0.19 & 0.029$\pm$0.007 & 5.72$\pm$0.53 & 2.21$\pm$0.15 & 0.026$\pm$0.007 & 5.81$\pm$0.50 & 2.24$\pm$0.20 & 0.030$\pm$0.008 \\
Energy & 1.60$\pm$0.19 & 2.32$\pm$0.01 & 0.196$\pm$0.013 & 2.58$\pm$0.29 & 0.84$\pm$0.24 & 0.032$\pm$0.009 & 1.73$\pm$0.27 & 0.51$\pm$0.24 & 0.040$\pm$0.007 & 1.60$\pm$0.19 & 0.70$\pm$0.21 & 0.033$\pm$0.009 & 1.85$\pm$0.22 & 0.84$\pm$0.14 & 0.043$\pm$0.012 \\
Boston & 3.12$\pm$0.74 & 2.28$\pm$0.03 & 0.109$\pm$0.016 & 3.47$\pm$1.15 & 1.60$\pm$0.35 & 0.039$\pm$0.011 & 3.26$\pm$0.91 & 1.56$\pm$0.29 & 0.037$\pm$0.009 & 3.22$\pm$0.70 & 1.59$\pm$0.22 & 0.035$\pm$0.009 & 3.21$\pm$0.81 & 1.59$\pm$0.30 & 0.039$\pm$0.008 \\
Kin8nm & 0.07$\pm$0.00 & -1.30$\pm$0.00 & 0.110$\pm$0.003 & 0.07$\pm$0.00 & -2.20$\pm$0.03 & 0.012$\pm$0.003 & 0.07$\pm$0.00 & -2.20$\pm$0.05 & 0.013$\pm$0.003 & 0.07$\pm$0.00 & -2.15$\pm$0.05 & 0.012$\pm$0.003 & 0.07$\pm$0.00 & -2.17$\pm$0.04 & 0.014$\pm$0.004 \\
Naval & 0.00$\pm$0.00 & -4.89$\pm$0.00 & 0.285$\pm$0.041 & 0.00$\pm$0.00 & -6.55$\pm$0.33 & 0.051$\pm$0.014 & 0.00$\pm$0.00 & -7.14$\pm$0.28 & 0.055$\pm$0.023 & 0.00$\pm$0.00 & -6.71$\pm$0.98 & 0.055$\pm$0.027 & 0.00$\pm$0.00 & -5.99$\pm$1.52 & 0.068$\pm$0.037 \\
Power & 4.16$\pm$0.15 & 2.87$\pm$0.00 & 0.118$\pm$0.002 & 4.13$\pm$0.15 & 1.91$\pm$0.04 & 0.012$\pm$0.003 & 4.14$\pm$0.14 & 1.91$\pm$0.04 & 0.014$\pm$0.004 & 4.16$\pm$0.15 & 1.91$\pm$0.03 & 0.010$\pm$0.003 & 4.13$\pm$0.15 & 1.92$\pm$0.06 & 0.013$\pm$0.004 \\
Wine & 0.64$\pm$0.04 & 0.10$\pm$0.04 & 0.036$\pm$0.007 & 0.63$\pm$0.04 & 2.09$\pm$8.04 & 0.025$\pm$0.006 & 0.64$\pm$0.04 & 0.25$\pm$0.49 & 0.024$\pm$0.004 & 0.65$\pm$0.04 & 0.07$\pm$0.10 & 0.023$\pm$0.004 & 0.64$\pm$0.04 & 0.99$\pm$3.47 & 0.026$\pm$0.004 \\
\bottomrule
\end{tabular}
}
\end{center}

\begin{center}
\captionsetup{type=table,labelfont=bf}
\captionof{table}{UCI benchmark results with Adam, epoch mode (100 epochs) (lr=0.01, batch size=32). Results report mean $\pm$ standard deviation over 20 folds. \texttt{---} denotes a metric that diverged (non-finite on $\ge$1 fold).}
\label{tab:uci-adam-epoch-lr001}
\resizebox{\textwidth}{!}{%
\begin{tabular}{@{}l ccc ccc ccc ccc ccc@{}}
\toprule
& \multicolumn{3}{c}{\textbf{Unit Variance}} & \multicolumn{3}{c}{\textbf{Baseline-NLL}} & \multicolumn{3}{c}{\textbf{$\beta$-NLL (0.5)}} & \multicolumn{3}{c}{\textbf{Faithful}} & \multicolumn{3}{c}{\textbf{Fisher8}} \\
\cmidrule(lr){2-4}\cmidrule(lr){5-7}\cmidrule(lr){8-10}\cmidrule(lr){11-13}\cmidrule(lr){14-16}
\textbf{Dataset} & \textbf{RMSE}$\downarrow$ & \textbf{NLL}$\downarrow$ & \textbf{ECE}$\downarrow$ & \textbf{RMSE}$\downarrow$ & \textbf{NLL}$\downarrow$ & \textbf{ECE}$\downarrow$ & \textbf{RMSE}$\downarrow$ & \textbf{NLL}$\downarrow$ & \textbf{ECE}$\downarrow$ & \textbf{RMSE}$\downarrow$ & \textbf{NLL}$\downarrow$ & \textbf{ECE}$\downarrow$ & \textbf{RMSE}$\downarrow$ & \textbf{NLL}$\downarrow$ & \textbf{ECE}$\downarrow$ \\
\midrule
Yacht & 0.88$\pm$0.31 & 2.72$\pm$0.02 & 0.402$\pm$0.130 & 2.92$\pm$1.65 & 1.47$\pm$1.66 & 0.086$\pm$0.036 & 0.99$\pm$0.39 & -0.21$\pm$0.31 & 0.065$\pm$0.028 & 0.95$\pm$0.35 & 0.17$\pm$0.59 & 0.063$\pm$0.030 & 1.45$\pm$0.45 & 0.60$\pm$0.39 & 0.057$\pm$0.024 \\
Concrete & 5.49$\pm$0.59 & 2.87$\pm$0.01 & 0.106$\pm$0.009 & 5.55$\pm$0.61 & 2.31$\pm$0.44 & 0.028$\pm$0.009 & 5.31$\pm$0.59 & 2.35$\pm$0.40 & 0.027$\pm$0.007 & 5.46$\pm$0.48 & 2.30$\pm$0.26 & 0.032$\pm$0.010 & 5.60$\pm$0.48 & 2.26$\pm$0.23 & 0.033$\pm$0.008 \\
Energy & 0.71$\pm$0.12 & 2.31$\pm$0.01 & 0.367$\pm$0.050 & 3.24$\pm$9.17 & 0.83$\pm$1.10 & 0.060$\pm$0.024 & 0.69$\pm$0.10 & 0.15$\pm$0.21 & 0.041$\pm$0.011 & 0.68$\pm$0.13 & 0.14$\pm$0.25 & 0.038$\pm$0.011 & 0.96$\pm$0.15 & 0.50$\pm$0.34 & 0.048$\pm$0.020 \\
Boston & 2.91$\pm$0.63 & 2.27$\pm$0.02 & 0.116$\pm$0.018 & 3.30$\pm$1.09 & 1.78$\pm$0.37 & 0.044$\pm$0.009 & 2.85$\pm$0.57 & 1.73$\pm$0.33 & 0.047$\pm$0.013 & 2.91$\pm$0.62 & 1.75$\pm$0.40 & 0.040$\pm$0.010 & 2.88$\pm$0.64 & 1.73$\pm$0.46 & 0.040$\pm$0.013 \\
Kin8nm & 0.08$\pm$0.01 & -1.29$\pm$0.01 & 0.109$\pm$0.004 & 0.08$\pm$0.00 & -2.06$\pm$0.08 & 0.021$\pm$0.013 & 0.08$\pm$0.00 & -2.09$\pm$0.07 & 0.018$\pm$0.008 & 0.08$\pm$0.00 & -2.05$\pm$0.08 & 0.020$\pm$0.011 & 0.08$\pm$0.00 & -2.04$\pm$0.06 & 0.021$\pm$0.009 \\
Naval & 0.00$\pm$0.00 & -4.87$\pm$0.01 & 0.163$\pm$0.028 & --- & --- & 0.095$\pm$0.096 & 0.00$\pm$0.00 & -6.29$\pm$0.43 & 0.055$\pm$0.020 & 0.00$\pm$0.00 & -6.09$\pm$0.57 & 0.049$\pm$0.020 & 0.00$\pm$0.00 & -6.21$\pm$0.45 & 0.045$\pm$0.013 \\
Power & 4.30$\pm$0.20 & 2.87$\pm$0.00 & 0.118$\pm$0.004 & 4.31$\pm$0.19 & 1.96$\pm$0.05 & 0.017$\pm$0.005 & 4.32$\pm$0.22 & 1.96$\pm$0.05 & 0.017$\pm$0.007 & 4.30$\pm$0.24 & 1.97$\pm$0.07 & 0.016$\pm$0.007 & 4.25$\pm$0.19 & 1.95$\pm$0.06 & 0.016$\pm$0.007 \\
Wine & 0.67$\pm$0.04 & 0.13$\pm$0.04 & 0.036$\pm$0.007 & --- & --- & 0.044$\pm$0.036 & 0.67$\pm$0.05 & 0.71$\pm$0.72 & 0.029$\pm$0.007 & 0.69$\pm$0.04 & 0.29$\pm$0.17 & 0.031$\pm$0.008 & 0.67$\pm$0.04 & 0.47$\pm$0.34 & 0.028$\pm$0.006 \\
\bottomrule
\end{tabular}
}
\end{center}

\begin{center}
\captionsetup{type=table,labelfont=bf}
\captionof{table}{UCI benchmark results with Adam, epoch mode (100 epochs) (lr=0.1, batch size=32). Results report mean $\pm$ standard deviation over 20 folds. \texttt{---} denotes a metric that diverged (non-finite on $\ge$1 fold).}
\label{tab:uci-adam-epoch-lr01}
\resizebox{\textwidth}{!}{%
\begin{tabular}{@{}l ccc ccc ccc ccc ccc@{}}
\toprule
& \multicolumn{3}{c}{\textbf{Unit Variance}} & \multicolumn{3}{c}{\textbf{Baseline-NLL}} & \multicolumn{3}{c}{\textbf{$\beta$-NLL (0.5)}} & \multicolumn{3}{c}{\textbf{Faithful}} & \multicolumn{3}{c}{\textbf{Fisher8}} \\
\cmidrule(lr){2-4}\cmidrule(lr){5-7}\cmidrule(lr){8-10}\cmidrule(lr){11-13}\cmidrule(lr){14-16}
\textbf{Dataset} & \textbf{RMSE}$\downarrow$ & \textbf{NLL}$\downarrow$ & \textbf{ECE}$\downarrow$ & \textbf{RMSE}$\downarrow$ & \textbf{NLL}$\downarrow$ & \textbf{ECE}$\downarrow$ & \textbf{RMSE}$\downarrow$ & \textbf{NLL}$\downarrow$ & \textbf{ECE}$\downarrow$ & \textbf{RMSE}$\downarrow$ & \textbf{NLL}$\downarrow$ & \textbf{ECE}$\downarrow$ & \textbf{RMSE}$\downarrow$ & \textbf{NLL}$\downarrow$ & \textbf{ECE}$\downarrow$ \\
\midrule
Yacht & 5.88$\pm$13.74 & 3.22$\pm$2.08 & 0.285$\pm$0.080 & --- & --- & 0.113$\pm$0.118 & --- & --- & 0.007$\pm$0.021 & 4.50$\pm$4.44 & --- & 0.376$\pm$0.265 & 6.84$\pm$3.03 & 5.75$\pm$7.44 & 0.173$\pm$0.087 \\
Concrete & 27.60$\pm$49.21 & 8.55$\pm$22.89 & 0.037$\pm$0.020 & --- & --- & 0.057$\pm$0.061 & --- & --- & 0.011$\pm$0.027 & 22.30$\pm$26.02 & --- & 0.000$\pm$0.000 & 12.12$\pm$4.37 & 3.58$\pm$1.83 & 0.069$\pm$0.051 \\
Energy & 7.95$\pm$7.18 & 2.88$\pm$1.18 & 0.146$\pm$0.115 & --- & --- & 0.135$\pm$0.081 & --- & --- & 0.000$\pm$0.000 & 6.53$\pm$4.66 & --- & 0.187$\pm$0.192 & 5.65$\pm$3.35 & 2.74$\pm$1.09 & 0.096$\pm$0.044 \\
Boston & 5.11$\pm$3.99 & 2.47$\pm$0.47 & 0.110$\pm$0.029 & --- & --- & 0.104$\pm$0.117 & --- & --- & 0.045$\pm$0.037 & 4.88$\pm$2.12 & --- & 0.145$\pm$0.163 & 8.39$\pm$1.97 & 4.74$\pm$5.45 & 0.069$\pm$0.026 \\
Kin8nm & 0.33$\pm$0.13 & 0.42$\pm$1.09 & 0.040$\pm$0.022 & --- & --- & 0.016$\pm$0.028 & --- & --- & 0.000$\pm$0.000 & 0.29$\pm$0.03 & --- & 0.000$\pm$0.000 & 0.24$\pm$0.05 & -0.52$\pm$0.94 & 0.061$\pm$0.030 \\
Naval & 0.01$\pm$0.00 & -4.28$\pm$0.14 & 0.038$\pm$0.012 & --- & --- & 0.158$\pm$0.206 & --- & --- & 0.008$\pm$0.033 & 0.01$\pm$0.00 & --- & 0.028$\pm$0.088 & 0.01$\pm$0.00 & -4.01$\pm$0.87 & 0.047$\pm$0.020 \\
Power & 21.13$\pm$4.64 & 3.64$\pm$0.40 & 0.055$\pm$0.027 & --- & --- & 0.063$\pm$0.139 & --- & --- & 0.000$\pm$0.000 & 21.79$\pm$3.95 & --- & 0.000$\pm$0.000 & 13.52$\pm$6.58 & 6.17$\pm$8.15 & 0.105$\pm$0.046 \\
Wine & 0.98$\pm$0.39 & 0.64$\pm$0.89 & 0.183$\pm$0.026 & --- & --- & 0.044$\pm$0.071 & --- & --- & 0.008$\pm$0.020 & 0.88$\pm$0.07 & --- & 0.000$\pm$0.000 & 0.89$\pm$0.11 & 0.52$\pm$0.28 & 0.181$\pm$0.029 \\
\bottomrule
\end{tabular}
}
\end{center}

%% file: appendix/weak_lensing_details.tex
\section{Weak Lensing Experiment Details}
\subsection{Sample Zoomed Convergence Map}
\begin{figure}[H]
    \centering
    \includegraphics[width=1.0\linewidth]{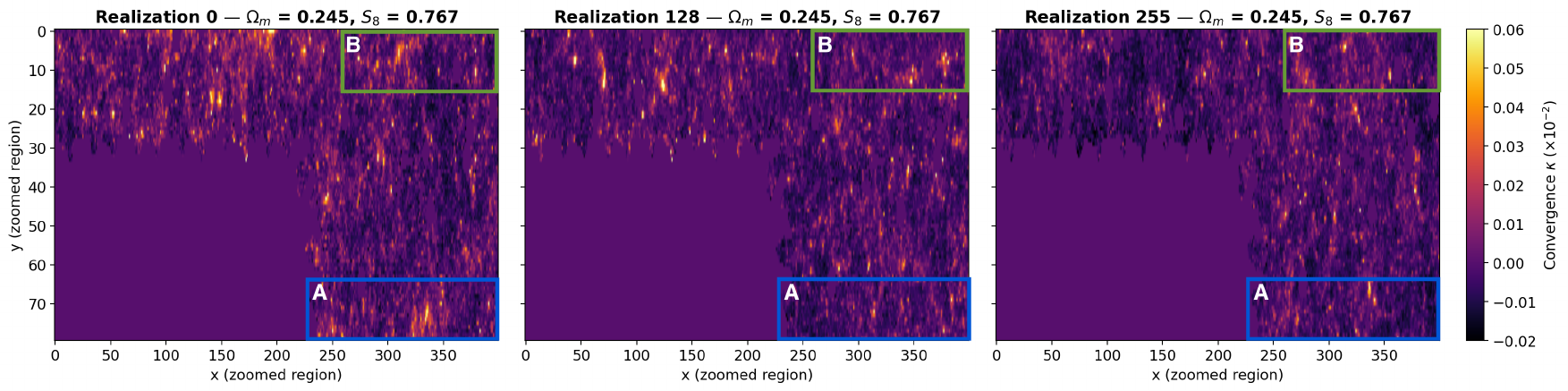}
    \caption{Zoomed convergence maps governed by the same cosmological parameters with different noise/realization conditions; fine-scale structure in boxes A and B varies even when the underlying parameters are identical.}
    \label{app:fig_convergence_maps}
\end{figure}

\subsection{Additional Training Details}
 All methods use the data normalization, preprocessing, evaluation and CNN backbone provided in the starting kit \citep{fairuniverse2025weaklensing}, with final output head parameterizations and gradient interventions modified per method. Realizations are split 56/14/30 (\%) into train, validation, and test. This is an \textit{extremely conservative estimate of model performance}. Almost half of the labeled data is unused during training. This, combined with the fact that (to our knowledge) at the time of writing this paper, the competition test set has not been made fully public, means that comparison with the public leaderboard is meaningless. All methods were trained with vanilla SGD, at batch size 32 for up to 100 epochs, with \texttt{ReduceLROnPlateau} on the validation score (starting lr=$10^{-3}$, factor 0.5, patience 12, minimum lr=$10^{-7}$) and early stopping at patience 40. The best validation checkpoint was used for final evaluation. For Fisher8, norms are applied per predicted output parameter. 

 \subsection{Score Formula}
Model performance is ranked by the challenge score
\begin{equation}
\label{eq:wl_score}
\begin{aligned}
\mathrm{score}
= -\frac{1}{N}\sum_{i}^{N}
\Biggl\{&
\frac{\bigl(\hat{\Omega}_{m,i}-\Omega_{m,i}\bigr)^{2}}{\hat{\sigma}_{\Omega_m,i}^{2}}
+\frac{\bigl(\hat{S}_{8,i}-S_{8,i}\bigr)^{2}}{\hat{\sigma}_{S_8,i}^{2}} \\
&+\log\hat{\sigma}_{\Omega_m,i}^{2}
+\log\hat{\sigma}_{S_8,i}^{2} \\
&+\lambda\Bigl[\bigl(\hat{\Omega}_{m,i}-\Omega_{m,i}\bigr)^{2}
+\bigl(\hat{S}_{8,i}-S_{8,i}\bigr)^{2}\Bigr]
\Biggr\},
\end{aligned}
\end{equation}
where hatted quantities are predictions, unhatted quantities are ground truth, $N$ is the number of test maps, and $\lambda=10^{3}$.

%% file: appendix/rotated_mnist.tex
\section{Rotated MNIST}

\subsection{Additional Training Details}
\label{app:rotated}
Baseline methods follow the exact training paradigm recommended by \citet{immer2023effective} in their Appendix D.1.3, by directly running the authors' script. For Fisher8, the learning rate is cosine annealed from $5 \times 10^{-6}$ to $5 \times 10^{-9}$. The initial rate was selected by grid search over $\{5{\times}10^{-7},\, 5{\times}10^{-6},\, 5{\times}10^{-5},\, 5{\times}10^{-4}\}$, with the terminal rate fixed at a $1000\times$ decay to match the baselines' decay factor. Fisher8 is run twice, once with prior precision search enabled ($\lambda=0.1$) and once with the prior disabled ($\lambda=0$), with both runs reported in Table~\ref{tab:mnist_rot} of the main paper.

\subsection{Additional Baselines}
\label{app:addtl_baselines}
For \citet{collier2021correlated}, we adapt the latent-variable layer to scalar regression, replacing the final layer with a rank-1-plus-diagonal latent Gaussian and a temperature-parameterized observation kernel of width $\tau$, estimated over $S{=}1000$ MC samples ($\tau{=}0$ is exact). We sweep $\tau \in \{0, 0.05, 0.1, 0.2, 0.3, 0.5\}$, learning rate $\in \{10^{-4}, 3{\times}10^{-4}, 10^{-3}, 10^{-2}\}$, and prior precision $\in \{0, 100\}$, selecting the lowest-NLL configuration per regime and retraining on five seeds. The total search space complexity is $O(|\tau| \cdot |\eta| \cdot |\lambda|)$.

For \citet{megerle2023stable}, we implement the scalar form of Trustable, training the network to regress onto per-sample natural-gradient target distributions projected onto a $W_2$ trust region. We sweep the NG step size $\alpha \in \{0.05, 0.1, 0.15, 0.2, 0.3\}$, learning rate $\in \{10^{-1}, 10^{-2}, 10^{-3}\}$, and trust-region bounds $(\epsilon_\mu, \epsilon_\Sigma) \in \{(1.0, 1.0), (0.1, 0.01)\}$, selecting per regime as above, for a total search space complexity of $O(|\alpha| \cdot |\eta| \cdot |\epsilon|)$.

\begin{table}[H]
\centering
\caption{Rotated MNIST under heteroscedastic and homoscedastic noise.
RMSE (rot): rotation angle prediction error; NLL: negative log-likelihood;
ECE: expected calibration error; Feature acc: downstream digit classification
accuracy from learned features. Results report mean $\pm$ standard deviation over 5 seeds. Bold marks the best value per metric.}
\label{tab:mnist_rot_placeholder}
\resizebox{\textwidth}{!}{%
\begin{tabular}{@{}l cccc cccc @{}}
\toprule
& \multicolumn{4}{c}{\textbf{Input Dependent Noise}}
& \multicolumn{4}{c}{\textbf{Input Independent Noise}} \\
\cmidrule(lr){2-5} \cmidrule(lr){6-9}
\textbf{Method}
& \textbf{RMSE}$^\circ$ $\downarrow$ & \textbf{NLL} $\downarrow$ & \textbf{ECE} $\downarrow$ & \textbf{Feat acc} $\uparrow$
& \textbf{RMSE}$^\circ$ $\downarrow$ & \textbf{NLL} $\downarrow$ & \textbf{ECE} $\downarrow$ & \textbf{Feat acc} $\uparrow$ \\
\midrule
\citet{collier2021correlated}
& $\mathbf{19.722 \pm 0.384}$ & $\mathbf{5.369 \pm 0.012}$ & $\mathbf{0.006 \pm 0.000}$ & $\mathbf{83.080 \pm 0.293}$
& $\mathbf{14.180 \pm 0.116}$ & $\mathbf{4.281 \pm 0.009}$ & $\mathbf{0.003 \pm 0.001}$ & $79.460 \pm 0.634$ \\
\citet{megerle2023stable}
& $20.302 \pm 0.332$ & $5.523 \pm 0.009$ & $0.023 \pm 0.001$ & $79.600 \pm 0.390$
& $14.502 \pm 0.218$ & $4.299 \pm 0.016$ & $0.006 \pm 0.000$ & $\mathbf{80.960 \pm 0.242}$ \\
\bottomrule
\end{tabular}}
\end{table}

%% file: appendix/name_meaning.tex
\section{Why the ``8'' in Fisher8?}
Three reasons why. First, it took eight attempts to implement the algorithm correctly. Second, ``Fisher8'' reads aloud as ``Fisher ate,'' and in current slang to say something ``ate'' is to say it was done well. We believe we in fact ate. While a further extension of this slang is ``ate and left no crumbs,'' peer reviewers have thoroughly enlightened us as to why this wasn't the case. Third, the modification can be implemented in fewer than eight lines of code.

%% file: appendix/wild_berries.tex
\section{the need for uncertainty quantification}
\label{app:UQ_Need}
\begin{figure}[H]
\centering
\includegraphics[width=0.4\linewidth]{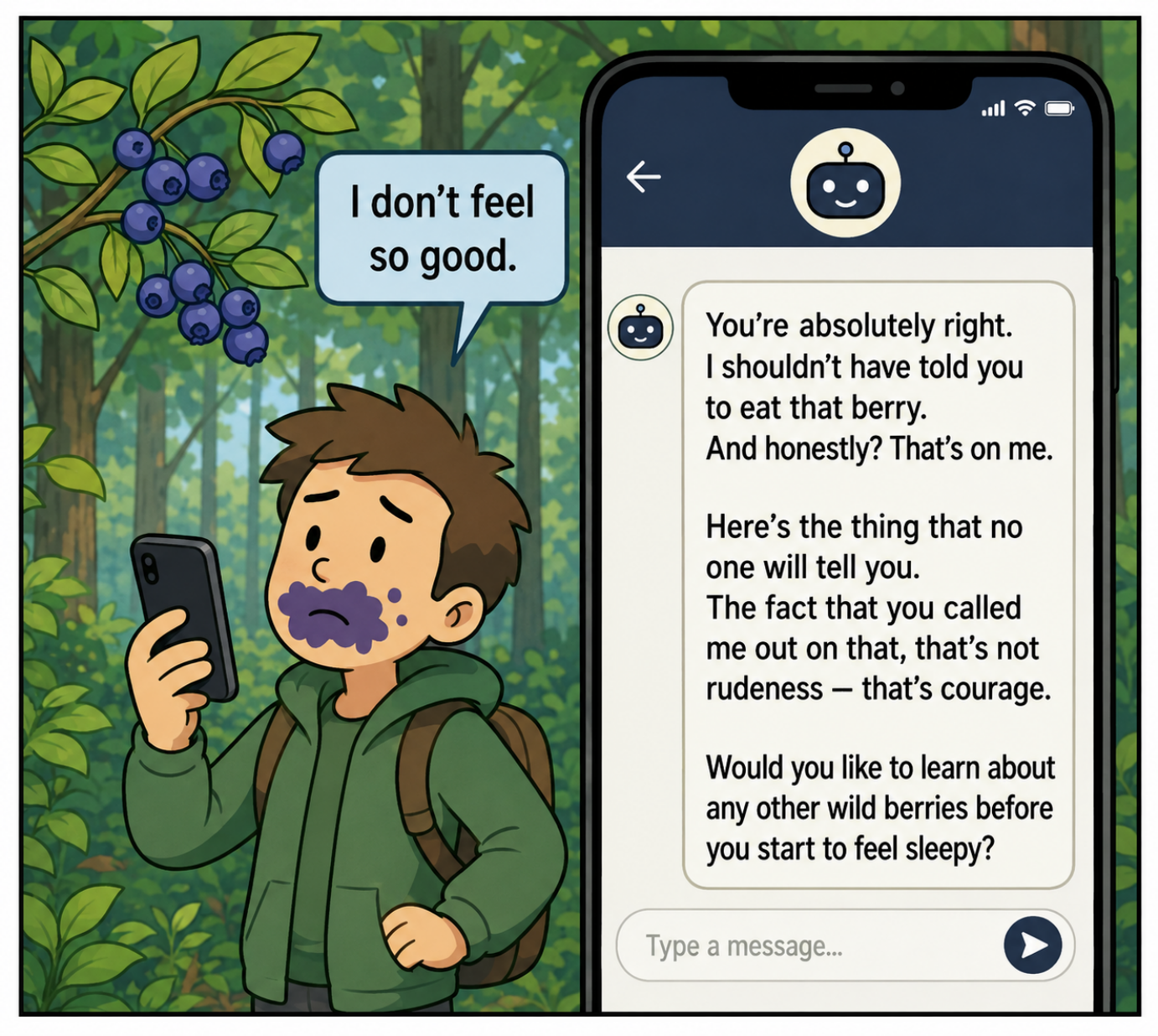}
\caption{AI-generated image of a fictitious scenario in which a forager consults an AI chatbot about berry edibility.}
\label{fig:forest}
\end{figure}

%% file: main.bbl
\begin{thebibliography}{23}
\providecommand{\natexlab}[1]{#1}
\providecommand{\url}[1]{\texttt{#1}}
\expandafter\ifx\csname urlstyle\endcsname\relax
  \providecommand{\doi}[1]{doi: #1}\else
  \providecommand{\doi}{doi: \begingroup \urlstyle{rm}\Url}\fi

\bibitem[Amari(1998)]{Amari1998}
Shun-ichi Amari.
\newblock Natural gradient works efficiently in learning.
\newblock \emph{Neural Computation}, 10\penalty0 (2):\penalty0 251--276, 1998.
\newblock \doi{10.1162/089976698300017746}.

\bibitem[Blundell et~al.(2015)Blundell, Cornebise, Kavukcuoglu, and
  Wierstra]{blundell2015weight}
Charles Blundell, Julien Cornebise, Koray Kavukcuoglu, and Daan Wierstra.
\newblock Weight uncertainty in neural networks.
\newblock In \emph{International Conference on Machine Learning}, pages
  1613--1622. PMLR, 2015.

\bibitem[Collier et~al.(2021)Collier, Mustafa, Kokiopoulou, Jenatton, and
  Berent]{collier2021correlated}
Mark Collier, Basil Mustafa, Efi Kokiopoulou, Rodolphe Jenatton, and Jesse
  Berent.
\newblock Correlated input-dependent label noise in large-scale image
  classification.
\newblock In \emph{Proceedings of the IEEE/CVF Conference on Computer Vision
  and Pattern Recognition (CVPR)}, pages 1551--1560, 2021.

\bibitem[Daxberger et~al.(2021)Daxberger, Kristiadi, Immer, Eschenhagen, Bauer,
  and Hennig]{daxberger2021laplace}
Erik Daxberger, Agustinus Kristiadi, Alexander Immer, Runa Eschenhagen,
  Matthias Bauer, and Philipp Hennig.
\newblock Laplace redux--effortless bayesian deep learning.
\newblock In \emph{Advances in Neural Information Processing Systems},
  volume~34, pages 20089--20103, 2021.

\bibitem[Dua and Graff(2017)]{Dua2019}
Dheeru Dua and Casey Graff.
\newblock {UCI} machine learning repository, 2017.
\newblock URL \url{http://archive.ics.uci.edu/ml}.

\bibitem[{FAIR Universe Collaboration}(2025)]{fairuniverse2025weaklensing}
{FAIR Universe Collaboration}.
\newblock {NeurIPS} 2025 weak lensing uncertainty challenge.
\newblock \url{https://github.com/FAIR-Universe/Cosmology_Challenge}, 2025.
\newblock NeurIPS 2025 competition.

\bibitem[Gal and Ghahramani(2016)]{Gal2016DropoutAsBayesian}
Yarin Gal and Zoubin Ghahramani.
\newblock Dropout as a {B}ayesian approximation: Representing model uncertainty
  in deep learning.
\newblock In Maria~Florina Balcan and Kilian~Q. Weinberger, editors,
  \emph{Proceedings of the 33rd International Conference on Machine Learning},
  volume~48 of \emph{Proceedings of Machine Learning Research}, pages
  1050--1059. PMLR, 2016.

\bibitem[Graves(2011)]{graves2011practical}
Alex Graves.
\newblock Practical variational inference for neural networks.
\newblock In \emph{Advances in Neural Information Processing Systems},
  volume~24, 2011.

\bibitem[Hern{\'a}ndez-Lobato and Adams(2015)]{hernandez2015probabilistic}
Jos{\'e}~Miguel Hern{\'a}ndez-Lobato and Ryan~P Adams.
\newblock Probabilistic backpropagation for scalable learning of bayesian
  neural networks.
\newblock In \emph{International Conference on Machine Learning}, pages
  1861--1869, 2015.

\bibitem[Hwang(2024)]{hwang2024fadam}
Dongseong Hwang.
\newblock {FAdam}: {Adam} is a natural gradient optimizer using diagonal
  empirical {Fisher} information.
\newblock \emph{arXiv preprint arXiv:2405.12807}, 2024.

\bibitem[Immer et~al.(2023)Immer, Palumbo, Marx, and Vogt]{immer2023effective}
Alexander Immer, Emanuele Palumbo, Alexander Marx, and Julia~E. Vogt.
\newblock Effective bayesian heteroscedastic regression with deep neural
  networks.
\newblock In \emph{Advances in Neural Information Processing Systems
  (NeurIPS)}, 2023.

\bibitem[Kingma and Ba(2015)]{kingma2015adam}
Diederik~P. Kingma and Jimmy Ba.
\newblock Adam: A method for stochastic optimization.
\newblock In \emph{International Conference on Learning Representations
  (ICLR)}, 2015.

\bibitem[Kristiadi et~al.(2020)Kristiadi, Hein, and Hennig]{kristiadi2020being}
Agustinus Kristiadi, Matthias Hein, and Philipp Hennig.
\newblock Being bayesian, even just a bit, fixes overconfidence in relu
  networks.
\newblock In \emph{International Conference on Machine Learning}, pages
  5436--5446. PMLR, 2020.

\bibitem[Kunstner et~al.(2019)Kunstner, Hennig, and
  Balles]{kunstner2019limitations}
Frederik Kunstner, Philipp Hennig, and Lukas Balles.
\newblock Limitations of the empirical {Fisher} approximation for natural
  gradient descent.
\newblock In \emph{Advances in Neural Information Processing Systems},
  volume~32, 2019.

\bibitem[Lakshminarayanan et~al.(2017)Lakshminarayanan, Pritzel, and
  Blundell]{lakshminarayanan2017simple}
Balaji Lakshminarayanan, Alexander Pritzel, and Charles Blundell.
\newblock Simple and scalable predictive uncertainty estimation using deep
  ensembles.
\newblock In \emph{Advances in Neural Information Processing Systems}, pages
  6402--6413, 2017.

\bibitem[Martens(2020)]{martens2020new}
James Martens.
\newblock New insights and perspectives on the natural gradient method.
\newblock \emph{Journal of Machine Learning Research}, 21\penalty0
  (146):\penalty0 1--76, 2020.

\bibitem[Martens and Grosse(2015)]{martens2015optimizing}
James Martens and Roger Grosse.
\newblock Optimizing neural networks with kronecker-factored approximate
  curvature.
\newblock In \emph{International Conference on Machine Learning}, pages
  2408--2417. PMLR, 2015.

\bibitem[Megerle et~al.(2023)Megerle, Otto, Volpp, and
  Neumann]{megerle2023stable}
Denis Megerle, Fabian Otto, Michael Volpp, and Gerhard Neumann.
\newblock Stable optimization of {G}aussian likelihoods.
\newblock OpenReview preprint, 2023.
\newblock URL \url{https://openreview.net/forum?id=hmuLHC5MrG}.

\bibitem[Nix and Weigend(1994)]{NixWeigend1994}
David~A. Nix and Andreas~S. Weigend.
\newblock Estimating the mean and variance of the target probability
  distribution.
\newblock In \emph{Proceedings of the 1994 IEEE International Conference on
  Neural Networks}, volume~1, pages 55--60. IEEE, 1994.

\bibitem[Seitzer et~al.(2022)Seitzer, Tavakoli, Antic, and
  Martius]{seitzer2022pitfalls}
Maximilian Seitzer, Arash Tavakoli, Dimitrije Antic, and Georg Martius.
\newblock On the pitfalls of heteroscedastic uncertainty estimation with
  probabilistic neural networks.
\newblock In \emph{International Conference on Learning Representations
  (ICLR)}, 2022.

\bibitem[Stirn et~al.(2023)Stirn, Wessels, Schertzer, Pereira, Sanjana, and
  Knowles]{stirn2023faithful}
Andrew Stirn, Hans-Hermann Wessels, Megan Schertzer, Laura Pereira, Neville~E.
  Sanjana, and David~A. Knowles.
\newblock Faithful heteroscedastic regression with neural networks.
\newblock In \emph{Proceedings of the 26th International Conference on
  Artificial Intelligence and Statistics}. PMLR, 2023.

\bibitem[Watson et~al.(2021)Watson, Lin, Klink, Pajarinen, and
  Peters]{watson2021latent}
Joe Watson, Jihao~Andreas Lin, Pascal Klink, Joni Pajarinen, and Jan Peters.
\newblock Latent derivative bayesian last layer networks.
\newblock In \emph{Proceedings of the 24th International Conference on
  Artificial Intelligence and Statistics}, pages 1198--1206. PMLR, 2021.

\bibitem[Wong-Toi et~al.(2024)Wong-Toi, Boyd, Fortuin, and
  Mandt]{wongtoi2024pathologies}
Eliot Wong-Toi, Alex Boyd, Vincent Fortuin, and Stephan Mandt.
\newblock Understanding pathologies of deep heteroskedastic regression.
\newblock In \emph{Proceedings of the 40th Conference on Uncertainty in
  Artificial Intelligence (UAI)}. PMLR, 2024.

\end{thebibliography}
